\documentclass[10pt,twocolumn,letterpaper]{article}

\usepackage{cvpr}              

\usepackage[dvipsnames]{xcolor}

\definecolor{cvprblue}{rgb}{0.21,0.49,0.74}
\newtheorem{definition}{Definition}
\usepackage[pagebackref,breaklinks,colorlinks,citecolor=cvprblue]{hyperref}
\usepackage[ruled,linesnumbered,lined]{algorithm2e}
\usepackage{pifont}
\usepackage{bbding}
\usepackage{threeparttable}
\usepackage{booktabs}
\usepackage{multirow}
\usepackage{xfrac}
\usepackage{amsmath}
\usepackage{colortbl}
\usepackage{tikz}
\usepackage{hyperref}

\def\paperID{1665} 
\def\confName{CVPR}
\def\confYear{2024}

\title{FedBRB: An Effective Solution to the Small-to-Large Scenario in Device-Heterogeneity Federated Learning}

\author{Ziyue Xu, Mingfeng Xu, Tianchi Liao, Zibin Zheng, Chuan Chen*\\
Sun Yat-sen University\\
{\tt\small xuzy53@mail2.sysu.edu.cn, chenchuan@mail.sysu.edu.cn}
}

\begin{document}
\maketitle
\begin{abstract}
    Recently, the success of large models has demonstrated the importance of scaling up model size. This has spurred interest in exploring collaborative training of large-scale models from federated learning perspective. Due to computational constraints, many institutions struggle to train a large-scale model locally. Thus, training a larger global model using only smaller local models has become an important scenario (i.e., the \textbf{small-to-large scenario}). Although recent device-heterogeneity federated learning approaches have started to explore this area, they face limitations in fully covering the parameter space of the global model.
    In this paper, we propose a method called \textbf{FedBRB} (\underline{B}lock-wise \underline{R}olling and weighted \underline{B}roadcast) based on the block concept. FedBRB can uses small local models to train all blocks of the large global model, and broadcasts the trained parameters to the entire space for faster information interaction. Experiments demonstrate FedBRB yields substantial performance gains, achieving state-of-the-art results in this scenario. Moreover, FedBRB using only minimal local models can even surpass baselines using larger local models.
\end{abstract}
\section{Introduction}
\label{sec:intro}
The recent success of large models has underscored the benefits of scaling up model size, including improved generalization and enhanced learning capabilities \cite{ray2023chatgpt}.
This has driven ever-growing demands on large-scale models training under the requirements of privacy protection \cite{guo2023aigc}.
As a privacy-preserving technique, Federated learning \cite{mcmahan2017communication} has gained widespread attention due to its ability to collaboratively train models while keeping user data decentralized and secure \cite{aledhari2020federated}, with applications in areas such as medical image analysis \cite{sohan2023systematic}, predictive text input \cite{hard2018federated} and finance data modeling \cite{dash2022federated}.
Thus, many institutions have been inspired to collaboratively train large-scale models using federated learning \cite{fan2023fate}.

\begin{figure}[t]
  \centering
  \setlength{\abovecaptionskip}{0.cm}
  \setlength{\belowcaptionskip}{-10pt}
  \includegraphics[width=8cm]{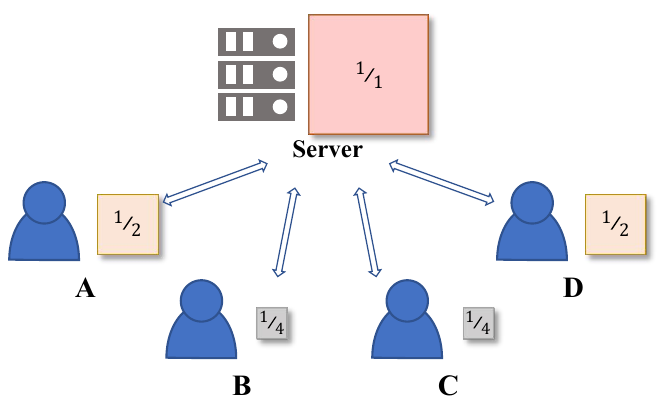}
  \caption{An illustration of the \textit{\textbf{small-to-large scenario}}, where the global model is larger than any client's local model. Squares of different sizes represent models of different sizes, where the fraction represents the model size.}
  \label{s2l}
\end{figure}

However, the limited computing capabilities of local devices in many institutions, including low-power terminal and mobile devices, often make performing complete large-scale local training infeasible.
This gives rise to the challenge of how to collaboratively train a model larger than any client's local model while preserving privacy and ensuring reasonable performance. In this paper we refer to the scenario for this challenge as the \textbf{\textit{small-to-large scenario}}, see \cref{s2l}.

Fortuitously, the paradigm of device-heterogeneity federated learning \cite{abdelmoniem2021towards} offers insights into this challenge.
Current device-heterogeneity federated learning aims at addressing device variations in federated learning by allowing devices with different computing capabilities, storage capacities, and network bandwidths to effectively participate in collaborative model training. To tackle device-heterogeneity, techniques based on knowledge distillation (KD) \cite{gou2021knowledge}, tensor decomposition (TD) \cite{9420085} and model partitioning (MP) \cite{diao2021heterofl, alam2022fedrolex} have been proposed. The common idea is to decompose a large size model into many small sub-models customized for different devices, thereby reducing the computation, memory, and communication costs.

KD-based methods \cite{cho2022heterogeneous, he2020group, itahara2021distillation, lin2020ensemble} often rely on public datasets to achieve competitive performance, which is infeasible in many cases. Their teacher-student framework is also incompatible with secure aggregation protocols, making them susceptible to backdoor attacks \cite{alam2022fedrolex}.
TD-based methods \cite{park2023fedhm, niu2022federated, mei2022resource} preclude the need for public data, but their tensor calculations introduce greater computational complexity.
In contrast, MP-based methods have attracted widespread attention in recent years due to their efficiency, intuitiveness, and ease of implementation. MP-based methods mainly refer to the segmenting the global model width-wise per layer.
Federated Dropout \cite{caldas2018expanding} utilizes Dropout to randomly select neurons per layer, but randomness also limits its effectiveness under conditions of high data heterogeneity and few clients \cite{alam2022fedrolex}.
HeteroFL \cite{diao2021heterofl} and FjORD \cite{horvath2021fjord} select the neurons in a fixed order, leaving large untrained areas and lacking support for the small-to-large scenario.
FedRolex \cite{alam2022fedrolex} improves HeteroFL by rollingly selecting different neurons in each round.
While FedRolex can be used in small-to-large scenario, it does not fully solve the problem of incomplete model coverage. Moreover, its traversal speed on model parameters depends on the model size, which would result in a negligible traversal time.
These limitations indicate room for further improvement.

In this paper we propose FedBRB (Block-wise Rolling and weighted Broadcast) based on the concept of the \textit{block}. FedBRB trains every block of the global model using local small models, and broadcasts the trained parameters to the full large model.
The former enables full training coverage of the global model, while the latter fuses updates from varying sub-models to accelerate the overall training process.
We conduct extensive evaluations of FedBRB on vision tasks under resource-constrained settings, where neither client has access to the full dataset, nor possesses the complete model.
Experimental results demonstrate that FedBRB achieves significant performance improvements in this field, reaching state-of-the-art results. Moreover, FedBRB using only minimal local models can even surpass baselines using larger local models.
This is important for institutions with limited computational power to participate in federated learning.

Our contributions can be summarized as follows:
\begin{itemize}
  \item We clearly define the \textit{small-to-large} scenario and explain its important significance. We propose FedBRB to specifically handle this scenario, which outperforms previous state-of-the-art methods.
  \item The proposed FedBRB addresses the limitations of existing methods in small-to-large scenario including incomplete coverage and slow traversal of model parameters.
  \item FedBRB can improve the performance of the global model under both iid and non-iid data distributions. And FedBRB using the smallest local model can even surpass baseline using larger local models.
\end{itemize}

\begin{table}[ht]\label{t2}
  \centering
  \setlength{\abovecaptionskip}{5pt}
  \setlength{\belowcaptionskip}{-10pt}
  \scalebox{0.75}{
    \begin{threeparttable}
      \setlength{\tabcolsep}{1mm}{
        \begin{tabular}{cccccc}
          \toprule
          \textbf{}                             & \textbf{\begin{tabular}[c]{c}Method \end{tabular}}     & \textbf{\begin{tabular}[c]{c}Support\\ Small-to-\\Large\end{tabular}} & \textbf{\begin{tabular}[c]{c}Avoid \\Public\\ Datasets\end{tabular}} & \textbf{\begin{tabular}[c]{c}Avoid\\ SVD \\Steps\end{tabular}} & \textbf{\begin{tabular}[c]{c}Fully\\ Trained\end{tabular}} \\ \midrule

          FedDF \cite{lin2020ensemble}          & \multirow{3}{*}{\begin{tabular}[c]{c}KD \end{tabular}} & \multirow{3}{*}{\CheckmarkBold}                                       & \multirow{3}{*}{\XSolidBrush}                                        & \multirow{3}{*}{\CheckmarkBold}                                & \multirow{3}{*}{\textbf{-}}                                \\
          DS-FL \cite{itahara2021distillation}  &                                                        &                                                                       &                                                                      &                                                                &                                                            \\
          Fed-ET \cite{cho2022heterogeneous}    &                                                        &                                                                       &                                                                      &                                                                &                                                            \\ \midrule
          FedHM \cite{park2023fedhm}            & \multirow{3}{*}{\begin{tabular}[c]{c}TD \end{tabular}} & \XSolidBrush                                                          & \CheckmarkBold                                                       & \XSolidBrush                                                   & \multirow{3}{*}{\textbf{-}}                                \\
          PriSM \cite{niu2022federated}         &                                                        & \CheckmarkBold                                                        & \CheckmarkBold                                                       & \XSolidBrush                                                   &                                                            \\
          FLANC \cite{mei2022resource}          &                                                        & \XSolidBrush                                                          & \CheckmarkBold                                                       & \CheckmarkBold                                                 &                                                            \\ \midrule
          FedDropout \cite{caldas2018expanding} & Random MP                                              & \XSolidBrush                                                          & \CheckmarkBold                                                       & \CheckmarkBold                                                 & \XSolidBrush                                               \\
          HeteroFL \cite{diao2021heterofl}      & Fixed MP                                               & \XSolidBrush                                                          & \CheckmarkBold                                                       & \CheckmarkBold                                                 & \XSolidBrush                                               \\
          FedRolex \cite{alam2022fedrolex}      & Rolling MP                                             & \CheckmarkBold                                                        & \CheckmarkBold                                                       & \CheckmarkBold                                                 & \XSolidBrush                                               \\
          \textbf{FedBRB}                       & BRB MP                                                 & \textbf{\CheckmarkBold}                                               & \textbf{\CheckmarkBold}                                              & \textbf{\CheckmarkBold}                                        & \textbf{\CheckmarkBold}                                    \\ \bottomrule
        \end{tabular}
      }
      \begin{tablenotes}
        \item \textit{MP means Model Partitioning.}
      \end{tablenotes}
    \end{threeparttable}
  }
  \caption{A comparison of FedBRB against the latest device-heterogeneity federated learning methods.}
\end{table}

\section{Related Work}
\paragraph{Heterogeneity Federated Learning.}
Federated Learning (FL) is a privacy-preserving training framework. FedAvg \cite{mcmahan2017communication} is the \textit{de facto} algorithm for FL.
Since FedAVG, various models \cite{zhang2021survey} have emerged that aim to address heterogeneity issue \cite{singh2022federated} caused by differences among clients in such as data \cite{zhao2018federated} and device \cite{diao2021heterofl}.
\underline{Data-heterogeneity}, also known as non-iid data, is very common in federated learning such as different data quantity and different label distribution \cite{zhao2018federated}.
The formal leads to the aggregated model being biased towards the sub-models with more data, while the latter negatively impacts clients with missing labels. FedProx \cite{li2020federated} adds a regularization term to penalize parameter differences between clients. MOON \cite{li2021model} adopts contrastive loss to improve the representation learning. In addition to model improvements, solutions have explored client grouping \cite{chen2022cfl, wolfrath2022haccs} and data processing.
\underline{Device-heterogeneity FL} \cite{abdelmoniem2021towards} methods can be divided into three branches including knowledge distillation, tensor decomposition and model partitioning, which we will introduce in detail next.
\vspace{-12pt}

\paragraph{Knowledge Distillation.}
FedDF \cite{lin2020ensemble} transfers knowledge from a group of classifier models trained with private data to the student model using teacher-student framework. DS-FL \cite{itahara2021distillation} further improves the effectiveness by incorporating an unlabeled dataset at the server and utilizing pseudo-labeling techniques. Fed-ET \cite{cho2022heterogeneous} weighted consensus distillation scheme with diversity regularization to extract reliable consensus and improve generalization.
However, methods based on knowledge distillation usually require a unrealistic well-crafted public dataset, and the teacher-student framework make them susceptible to backdoor attacks \cite{wang2020attack}.
\vspace{-12pt}

\paragraph{Tensor Decomposition.}
FedHM \cite{park2023fedhm} uses truncated SVD \cite{golub1971singular} to decompose a 2D convolution filter into two 1-D filters to reduce the memory cost. However, FedHM still faces challenges in peak memory usage \cite{niu2022federated}.
PriSM \cite{niu2022federated} maps convolutional kernels into an orthogonal space using SVD to obtain the principal orthogonal kernels, and sample different subsets of orthogonal kernels to create sub-models.
However, the computational overhead introduced by using SVD is non-negligible.
Unlike decomposition approaches, FLANC \cite{mei2022resource} utilizes a neural composition method to adaptively construct sub-models of appropriate sizes based on a globally shared tensor. Although FLANC enables the global flow of information and avoids SVD step, it cannot obtain a global model larger than the largest client model.
\vspace{-12pt}

\paragraph{Model Partitioning.}
Federated Dropout \cite{caldas2018expanding} obtains sub-models by randomly selecting neurons in each layer.
But randomness also limits its effectiveness.
HeteroFL \cite{diao2021heterofl} and FjORD \cite{horvath2021fjord} extracts neurons in a fixed order so that early neurons learn general patterns while later ones capture abstract knowledge.Unfortunately, fixed selection is unsuitable for the small-to-large scenario.
FedRolex \cite{alam2022fedrolex} enhances upon HeteroFL by modifying the selected neurons in each round using a method similar to sliding window.
Although can be utilized in the small-to-large scenario, large portions of FedRolex's parameters remain untrained, underscoring the need to investigate a better extraction method.

\section{Method}
In this section, we first define the small-to-large scenario. Then we provide a detailed analysis of the current model partitioning methods, and offer a detailed and visual description of our proposed FedBRB.

\vspace{-12pt}
\paragraph{Notation.}
For a classification task, the total training data set is denoted as $D$, which consists of data $X$ and labels $Y$.
The clients in federated learning are denoted as $C$, and the total number of clients is $N_C$. An individual client is denoted as $C_i$ based on its index.
$C_i$'s local model is parameterized by model parameters $W_i$.
The participating clients upload their updated model parameters, and the server aggregates these parameters to obtain the global model $W_g$.
\vspace{-12pt}
\begin{definition}
  (\textbf{The Small-to-Large Scenario}) is an extreme device-heterogeneity FL scenario where the size of the local models on each client is \textbf{strictly smaller} than the size of the global model on server:
\end{definition}
\begin{equation}
  \exists d, \forall i, \forall t: \operatorname{dim}_d(\textbf{W}_i^{t}) < \operatorname{dim}_d( \textbf{W}{g}^{t}),
  \label{definition_equation}
\end{equation}
where $i$ is client index, $W^t$ is the $t$-th tensor of $W$, and $\operatorname{dim}_d(\cdot)$ is getting the size of $d$-th dimension of the given tensor.
Specifically, \cref{definition_equation} means that there exists some dimension in each layer of each sub-model whose size is strictly smaller than the size of that dimension of the corresponding tensor in the global model network. As mainstream model sizes increase, this scenario will become increasingly common and important.

\vspace{-5pt}
\begin{figure}[h]
  \centering
  \setlength{\abovecaptionskip}{0.cm}
  \setlength{\belowcaptionskip}{-10pt}
  \includegraphics[width=8cm]{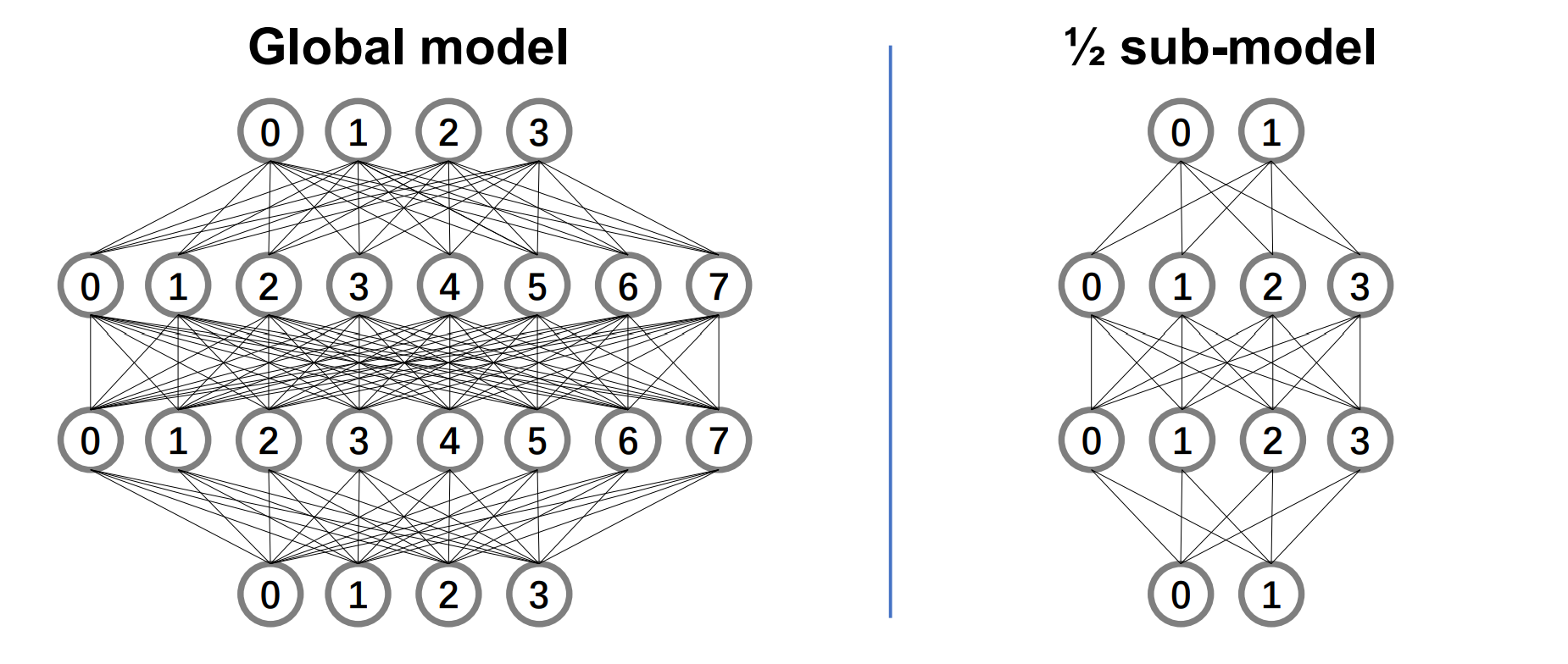}
  \caption{\label{fig_2} Model partitioning in neuron perspective.}
\end{figure}

\subsection{Model Partitioning}
Model partitioning methods obtain sub-models by segmenting the global model width-wise per layer.
We explain this from two perspectives - the neuron perspective and the parameter tensor perspective, as a single perspective can lead to some issues being glossed over.

\vspace{-12pt}
\paragraph{$\alpha$. Neuron Perspective}
Let $l$ be the $l$-th layer of a deep neural network, $\mathcal{N}^{l}$ be the neurons on $l$-th layer and $\mathcal{R}_i \in (0, 1]$ be the $i$-th client's partitioning ratio. Then the formal formula can be expressed as follows:
\begin{equation}
  \label{np}
  \mathcal{N}_{i}^{l}[j] = \mathcal{N}_{g}^{l}[\chi^{l}_{i}][j],
\end{equation}
where $\mathcal{N}_{i}^{l}[j]$ denotes the $j$-th neuron on $l$-th layer of $i$-th client's sub-model, $g$ denotes the global model, and $\chi^{l}_{i}$ denotes the $i$-th client's selecting index sequence for the $l$-th layer. \cref{np} means selecting neurons from $\mathcal{N}_{g}^{l}$ according to the selecting sequence $\chi^{l}_{i}$.
Note $|\chi^{l}_{i}| \equiv |\mathcal{N}_{g}^{l}| \times \mathcal{R}^{l}_{i}$. \cref{fig_2} is an illustration of $\frac{1}{2}$ partitioning from neuronal perspective.

\begin{figure*}[t]
  \centering
  \setlength{\abovecaptionskip}{0.cm}
  \setlength{\belowcaptionskip}{-10pt}
  \includegraphics[width=17.5cm]{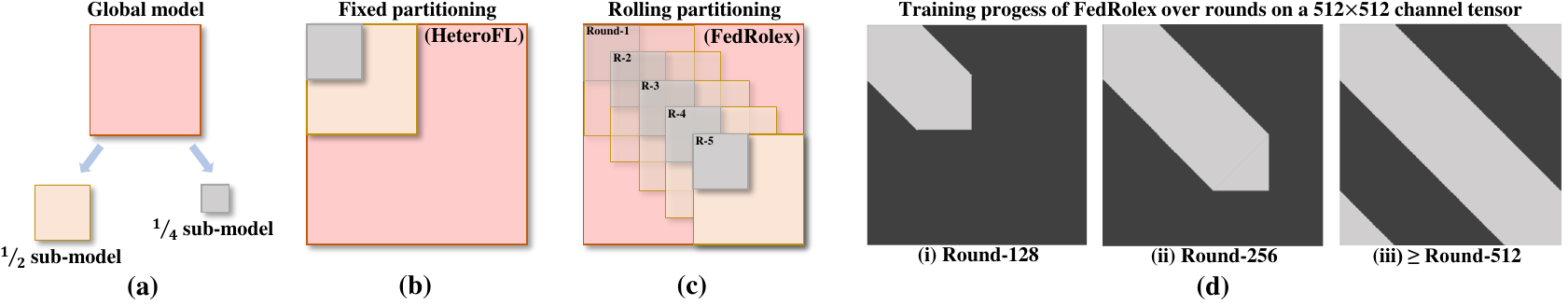}
  \caption{\label{fig_1} (a) Model partitioning illustration in parameter tensor perspective. (b) Fixed partitioning illustration where the trained area remains fixed over rounds. (c) Rolling partitioning illustration where the trained area changes over rounds. (d) An example showing that baseline FedRolex trains a $\sfrac{1}{4}$ sub-model on a 512$\times$512 channel tensor, with a large portion of the parameters remain untrained (black).}
\end{figure*}

\vspace{-12pt}
\paragraph{$\beta$. Parameter Tensor Perspective}
Specifically in the vision field, the kernels between the ($l$-1)-th and $l$-th convolution layers in the complete global model can be seen as a 4-$D$ tensor $W^{l-1:l} \in R^{M,N,k,k}$, where $M$ and $N$ respectively represent the number of output and input channels, and $k$ is the convolution kernel size. $N$ and $M$ are also the number of neurons in the ($l$-1)-th layer and the $l$-th layer, respectively.
Then the model parameters between the ($l$-1)-th and $l$-th layers of the $i$-th client's sub-model obtained by partitioning ratio $\mathcal{R}_{i}$ can be denoted as $W^{l-1:l}_{i} \in R^{M \times \mathcal{R}_{i}, N \times \mathcal{R}_{i}, k, k}$. It is achieved by picking the channels of input and output to obtain tensor slices and construct sub-models.
An illustration of parameter tensor perspective partitioning is shown in \cref{fig_1}(a).

From parameter tensor perspective, the essential differences among existing methods lies in the channel-selection sequence scheme.
The sequences between the ($l$-1)-th and $l$-th layers consists of two sequences - the input channel sequence and the output channel sequence. Federated Dropout uses random sequences to select input and output channels \cite{alam2022fedrolex}, while HeteroFL uses continuous sequences with fixed starting numbers \cite{diao2021heterofl}. FedRolex's rolling sequences for input and output channels are similar to HeteroFL's but differs in the increasing starting number over rounds.
Figure \cref{fig_1}(b) and \cref{fig_1}(c) illustrate HeteroFL and FedRolex from the parameter tensor perspective. The FedRolex increment interval we demonstrated in \cref{fig_1}(c) is relatively large, which is to improve the clarity of the demonstration. In fact, FedRolex only increments the input and output sequence starting numbers by 1 simultaneously per round, which is a very small increment interval.

\begin{table}[t]
  \centering
  \setlength{\abovecaptionskip}{5pt}
  \setlength{\belowcaptionskip}{-15pt}
  \scalebox{0.9}{
    \begin{tabular}{ccccc}
      \toprule
      \textbf{} & \textbf{conv1} & \textbf{conv2} & \textbf{conv3} & \textbf{conv4} \\ \midrule
      block1    & {[}64, 3{]}    & {[}64, 64{]}   & {[}64, 64{]}   & {[}64, 64{]}   \\
      block2    & {[}64, 128{]}  & {[}128, 128{]} & {[}128, 128{]} & {[}128, 128{]} \\
      block3    & {[}128, 256{]} & {[}256, 256{]} & {[}256, 256{]} & {[}256, 256{]} \\
      block4    & {[}256, 512{]} & {[}512, 512{]} & {[}512, 512{]} & {[}512, 512{]} \\ \bottomrule
    \end{tabular}
  }
  \caption{\label{t1} The output-input channel numbers of conv-tensor in ResNet18 . The size of kernel is hidden which is 3 here, e.g. [64, 3, 3, 3] of block1-conv1.}
\end{table}

FedRolex uses neuron perspective to illustrate its model and claims that it can train all neurons evenly. 
However, in practice, the parameters of neural networks are not the neurons themselves, but rather the connection weights between neurons across layers.
If we analyze FedRolex from parameter tensor perspective, we can find that it still has two shortcomings: (1) Slow rolling speed, and (2) A large portion of parameter space remains untrained.
To illustrate in detail, we analyze the most commonly used ResNet18 as an example, whose architecture is shown in \cref{t1}.
First, the maximum number of channels for the convolutional kernel in ResNet18 is 512, hence FedRolex necessitates 512 training rounds to complete a single traversal, which means unrealistic communication costs in federated learning. Second, as there are lots of tensors in ResNet18 with equal input and output channels, the selected sequence of input and output channels in FedRolex would be the same. Such sequences of input and output will result in a cyclic traversal of the diagonal slices of the kernel tensor, as exemplified in \cref{fig_1}(d), which describes the traversal process over a 512-512 channel tensor.
The black areas represent parts untrained up to a given round, while the white areas denote regions trained at least once. When the currently trained slice of FedRolex reaches the lower right edge, it starts over from the beginning due to the square shape of the tensor, leaving black areas untrained regardless of how many rounds pass.

\begin{figure*}[t]
  \centering
  \setlength{\abovecaptionskip}{0.cm}
  \setlength{\belowcaptionskip}{-10pt}
  \includegraphics[width=17cm]{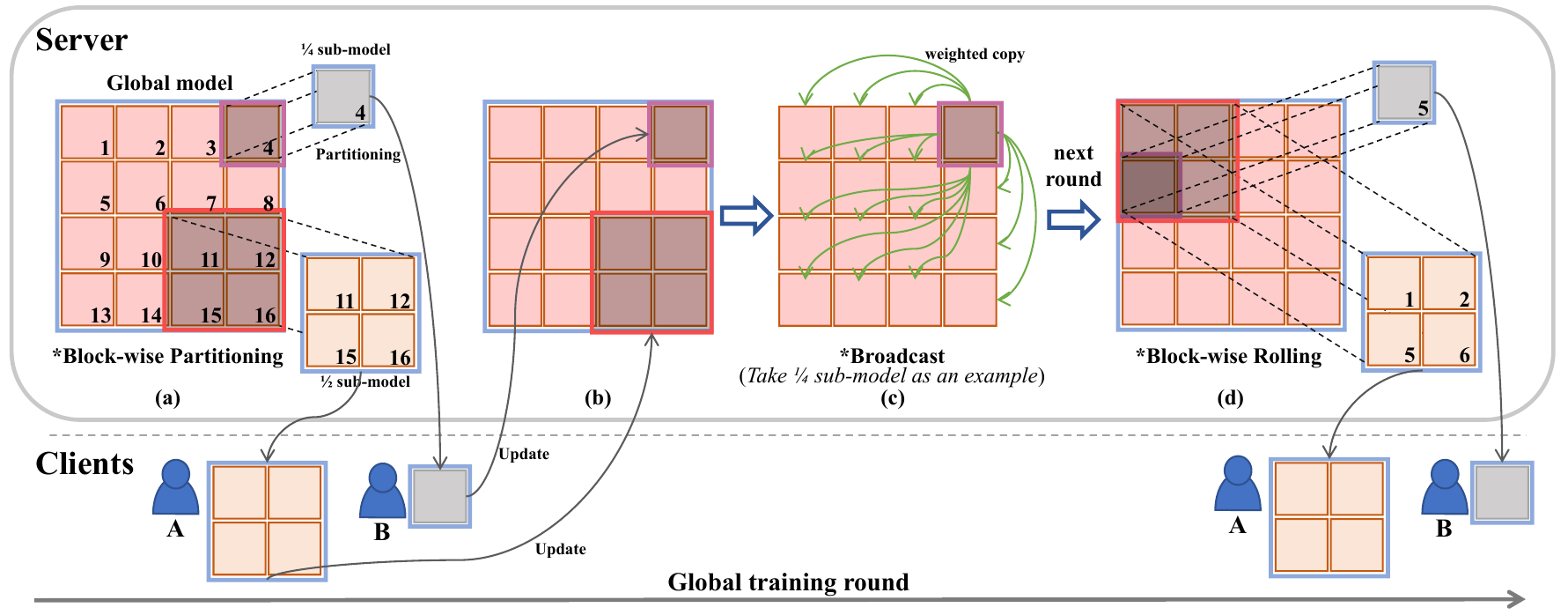}
  \caption{\label{FedBRB_fig} Overview of the FedBRB. (a) The global model is partitioned into 16 blocks based on the size of the smallest client B. Client B uses 1 block as its sub-model, while client A combines 4 blocks as its sub-model. (b) Clients upload their updates. (c) Weighted broadcasting shares the updated blocks to all other places to further increases the frequency of information interaction between sub-models. (d) After server aggregation, the partitioning place of blocks will roll to the next index, enabling FedBRB to train all parameters of the global model over successive rounds.}
\end{figure*}

\subsection{The Proposed FedBRB}
To address the aforementioned limitations of FedRolex, we propose \textbf{FedBRB}: \textbf{B}lock-wise \textbf{R}olling and block weighted \textbf{B}roadcast.
The core methodology of \textit{Block-wise Rolling} is to partition the global model into blocks and conduct global rolling training in a block-wise manner, which provides two benefits: (1) enabling training to fully cover global model parameters, and (2) accelerating training traversal over parameters. And the module \textit{block weighted broadcast} refers to weightedly copy the updated block to all other blocks which accelerates information interaction between different sub-models.

\vspace{-5pt}
\subsubsection{Block-wise Partitioning and Rolling.}
\begin{definition}
  (\textbf{Block}) is a slice of the global model parameter tensors in a given layer. The shape of a block is identical to the smallest local sub-model tensor in that layer.
\end{definition}
\vspace{-5pt}

We formalize the block based on network layer tensors. Suppose a conv-tensor $T$ is of size $[M,N,k,k]$, and the partitioning ratios of clients are $\{R\}$. These ratios refer to the input and output channel selection proportions in each global network layer, and are typically reciprocal powers of 2 for compatibility, as described in HeteroFL \cite{diao2021heterofl}. Denote the consequent slices of $T$ from $s$-th to ($e$-1)-th slice in the 1st dimension as $T\textbf{[}s$:$e,:,:,:\textbf{]}$, and the input and output channel size $|B_{out}|$ and $|B_{in}|$ of the block can be formalized as:
\begin{equation}
  \label{block_channel_size_in}
  |B_{out}| = M \times \operatorname{min}(\{R\}), |B_{in}| = N \times \operatorname{min}(\{R\}).
\end{equation}
Therefore, the size of the block is $[|B_{out}|, |B_{in}|, k, k]$. We take this shape as the unit for partitioning and can neatly cut $T$ into $\frac{M}{|B_{out}|}\cdot\frac{N}{|B_{in}|}$ blocks. The specific formula for the \underline{$i$}-th block tensor is as follows:

\vspace{-12pt}
\begin{equation}
  \label{block}
  \begin{split}
    B_i & = T\textbf{[}(x-1)\times |B_{out}| : x\times |B_{out}|,            \\
        & \quad\quad(y-1)\times |B_{in}| : y\times |B_{in}|, :, :\textbf{]},
  \end{split}
\end{equation}
where $x$=$\lceil i \div \frac{N}{|B_{in}|}\rceil$ and $y$=$(i-1) \% \frac{N}{|B_{in}|}+1$.

\cref{FedBRB_fig}(a) shows the concept of the block.
Before training begins, each participating client selects a suitable size for its sub-model based on the their capabilities.
There are client A and B in this system, where A can handle $\sfrac{1}{2}$ of the input and output channels and B can only hold $\sfrac{1}{4}$. FedBRB determines the block size based on the smallest sub-models size. Thus the conv-tensor of each layer in global network will be evenly divided into 16 blocks since the smallest sub-model in this system is of size $\sfrac{1}{4}$.
FedBRB then takes the block as the minimum unit and assemble different sub-models with blocks for different sized clients based on their requirements (\eg, client A's sub-model is going to be assembled from 11-th, 12-th, 15-th and 16-th blocks).


In each global round, the server selects a starting block for every given model-size, and slices a correspondingly sized sub-model consisting of several blocks from that position, then distributes this sub-model to the corresponding clients.
After completing local training, the clients upload their updated blocks, and the server aggregates the received by overlaying them back onto the same positions from which they are sliced.
Furthermore, we propose the \textit{Block Weighted Broadcast} module during the aggregation, the details of which are deferred to a later subsection.

\begin{table*}[h]
  \centering
  \scalebox{0.8}{
    \begin{tabular}{cccccccccc}
      \toprule
      \textbf{CIFAR-10}       & \multicolumn{3}{c}{\textbf{non-iid-2}} & \multicolumn{3}{c}{\textbf{non-iid-5}} & \multicolumn{3}{c}{\textbf{iid}}                                                                                                                                                                                                                                           \\
      \cmidrule(r){1-1} \cmidrule(r){2-4}  \cmidrule(r){5-7} \cmidrule(r){8-10}
      \textbf{dynamic}        & \textbf{HeteroFL}                      & \textbf{FedRolex}                      & {\underline{\textbf{FedBRB}}}                & \textbf{HeteroFL}             & \textbf{FedRolex}             & {\underline{\textbf{FedBRB}}}                & \textbf{HeteroFL}             & \textbf{FedRolex}             & {\underline{\textbf{FedBRB}}}                \\ \midrule
      \textbf{a0-e1}          & \cellcolor[HTML]{C0C0C0}19.38          & \cellcolor[HTML]{C0C0C0}38.05          & \textbf{\underline{53.51}} ($^\uparrow$15.5) & \cellcolor[HTML]{C0C0C0}25.88 & \cellcolor[HTML]{C0C0C0}44.46 & \textbf{\underline{62.75}} ($^\uparrow$19.2) & \cellcolor[HTML]{C0C0C0}24.20 & \cellcolor[HTML]{C0C0C0}52.67 & \textbf{\underline{61.62}} ($^\uparrow$8.95) \\
      \textbf{a0-d1}          & \cellcolor[HTML]{C0C0C0}22.82          & \cellcolor[HTML]{C0C0C0}47.15          & \textbf{59.37} ($^\uparrow$12.2)             & \cellcolor[HTML]{C0C0C0}27.81 & 63.09                         & \textbf{65.52} ($^\uparrow$2.96)             & \cellcolor[HTML]{C0C0C0}25.41 & \textbf{70.67}                & 67.96 ($^\downarrow$2.71)                    \\
      \textbf{a0-c1}          & \cellcolor[HTML]{C0C0C0}23.57          & 55.73                                  & \textbf{68.87} ($^\uparrow$13.1)             & \cellcolor[HTML]{C0C0C0}29.92 & 71.51                         & \textbf{77.37} ($^\uparrow$6.05)             & \cellcolor[HTML]{C0C0C0}28.26 & 75.94                         & \textbf{78.12} ($^\uparrow$2.18)             \\
      \textbf{a0-b1}          & \cellcolor[HTML]{C0C0C0}30.52          & 65.41                                  & \textbf{73.72} ($^\uparrow$8.31)             & \cellcolor[HTML]{C0C0C0}42.30 & 81.87                         & \textbf{87.80} ($^\uparrow$5.93)             & \cellcolor[HTML]{C0C0C0}40.27 & 85.30                         & \textbf{90.04} ($^\uparrow$4.74)             \\ \midrule
      \textbf{a0-d1-e1}       & \cellcolor[HTML]{C0C0C0}25.69          & \cellcolor[HTML]{C0C0C0}48.29          & \textbf{55.89} ($^\uparrow$7.60)             & \cellcolor[HTML]{C0C0C0}37.60 & \cellcolor[HTML]{C0C0C0}61.16 & \textbf{66.92} ($^\uparrow$5.76)             & \cellcolor[HTML]{C0C0C0}30.61 & \textbf{68.90}                & 66.28 ($^\downarrow$2.62)                    \\
      \textbf{a0-c1-e1}       & \cellcolor[HTML]{C0C0C0}30.53          & \cellcolor[HTML]{C0C0C0}52.93          & \textbf{66.33} ($^\uparrow$13.4)             & \cellcolor[HTML]{C0C0C0}39.62 & 72.10                         & \textbf{78.92} ($^\uparrow$6.82)             & \cellcolor[HTML]{C0C0C0}34.68 & \textbf{74.74}                & 73.21 ($^\downarrow$1.53)                    \\
      \textbf{a0-b1-e1}       & \cellcolor[HTML]{C0C0C0}31.52          & 57.26                                  & \textbf{71.01} ($^\uparrow$13.8)             & \cellcolor[HTML]{C0C0C0}45.42 & 78.87                         & \textbf{84.49} ($^\uparrow$5.62)             & \cellcolor[HTML]{C0C0C0}46.19 & 84.52                         & \textbf{84.77} ($^\uparrow$0.25)             \\
      \textbf{a0-c1-d1}       & \cellcolor[HTML]{C0C0C0}30.51          & 56.39                                  & \textbf{67.01} ($^\uparrow$10.6)             & \cellcolor[HTML]{C0C0C0}37.23 & 72.52                         & \textbf{75.29} ($^\uparrow$2.77)             & \cellcolor[HTML]{C0C0C0}32.93 & 76.05                         & \textbf{76.07} ($^\uparrow$0.02)             \\
      \textbf{a0-b1-d1}       & \cellcolor[HTML]{C0C0C0}31.75          & 65.13                                  & \textbf{70.29} ($^\uparrow$5.16)             & \cellcolor[HTML]{C0C0C0}47.46 & 80.12                         & \textbf{83.35} ($^\uparrow$3.23)             & \cellcolor[HTML]{C0C0C0}45.57 & \textbf{83.65}                & 83.15 ($^\downarrow$0.50)                    \\
      \textbf{a0-b1-c1}       & \cellcolor[HTML]{C0C0C0}30.70          & 68.07                                  & \textbf{73.21} ($^\uparrow$5.14)             & \cellcolor[HTML]{C0C0C0}48.46 & 80.74                         & \textbf{84.81} ($^\uparrow$4.07)             & \cellcolor[HTML]{C0C0C0}48.37 & 84.55                         & \textbf{85.87} ($^\uparrow$1.32)             \\ \midrule
      \textbf{a0-c1-d1-e1}    & \cellcolor[HTML]{C0C0C0}27.34          & \cellcolor[HTML]{C0C0C0}47.36          & \textbf{67.40} ($^\uparrow$20.0)             & \cellcolor[HTML]{C0C0C0}38.25 & 71.34                         & \textbf{71.70} ($^\uparrow$0.36)             & \cellcolor[HTML]{C0C0C0}48.23 & 74.25                         & \textbf{78.17} ($^\uparrow$3.92)             \\ \midrule
      \textbf{a0-b1-c1-d1-e1} & \cellcolor[HTML]{C0C0C0}30.44          & 58.92                                  & \textbf{68.12} ($^\uparrow$9.20)             & \cellcolor[HTML]{C0C0C0}51.02 & 77.79                         & \textbf{82.31} ($^\uparrow$4.52)             & \cellcolor[HTML]{C0C0C0}54.38 & 81.60                         & \textbf{83.69} ($^\uparrow$2.9)              \\ \bottomrule
    \end{tabular}
  }
  \caption{(\textbf{Q1}) Global model accuracy comparison between FedBRB, HeteroFL and FedRolex on \textbf{CIFAR-10} under \textbf{dynamic} setting. The $^\uparrow$ indicates the improvement of FedBRB compared to FedRolex. The shadow area indicates the comparison method using larger clients which was surpassed by FedBRB using only the smallest size client (underlined).}
  \label{CIFAR10_dynamic}
\end{table*}

After aggregation, the server shifts the starting block forward, enabling FedBRB to traverse all parameters of the global model over consecutive rounds.
The starting blocks for different sized clients' sub-models are not the same, as we need to ensure that the assembled sub-model is consistent with the certain situation where the block of FedBRB is determined by the partitioning ratio of that client.
In \cref{FedBRB_fig}(a), client B receives a sub-model whose one layer is consisted of a single block sliced from the 4-th block. Then in next round, the server will assign client B the 5-th block, see \cref{FedBRB_fig}(d). Similarly, client A received blocks 11-th, 12-th, 15-th, and 16-th in the previous round, and it will receive blocks 1-st, 2-nd, 5-th, and 6-th in the next round.
Moreover, it can be observed from \cref{FedBRB_fig}(d) that sub-models of different sizes may sometimes overlap, which is a characteristic of FedBRB, enabling information interaction between different sub-models.

\vspace{-12pt}
\subsubsection{Block Weighted Broadcast.}
In pursuit of faster information interaction, we add a module named Block Weighted Broadcast during the aggregation, which not only applies updates to the originally sliced position, but also broadcasts them to all other block-regions of the global model. With such an approach, the information interaction between sub-models of different sizes occurs promptly in each round without waiting for the overlaps of sliced partitions which are usually quite rare.

See \cref{FedBRB_fig}(c) for illustration.
After receiving the $\sfrac{1}{4}$ sub-model sent from client B, the server will not only add its updated gradients to the 4-th block, but also copy them to the remaining 15 blocks with a predefined weight $\beta \in (0, 1)$. Similarly, the $\sfrac{1}{2}$ sub-model uploaded by A will be assigned to the 4 blocks in the lower right corner, and with weight $\beta$ to the upper left, lower left, and upper right 3 places.

\vspace{-12pt}
\subsubsection{Comparing to FedRolex.}
(1) \textbf{Trained areas}. FedBRB's block-wise rolling and broadcast provide full coverage of the global model parameters. In contrast, FedRolex's large parameters remain untrained. Traditional scenario covers up FedRolex's shortcomings because the client can have a complete model, while the small-to-large scenario fully expose its limitations.
(2) \textbf{Traversal speed}. Assuming that the smallest size in the system is $\sfrac{1}{n}$, FedBRB takes at most $n^2$ roundsto traverse once using block-wise rolling. While for FedRolex, assuming the tensor of a layer is of $[C_{in},C_{out},d,d]$ size, traversing the diagonal of that layer requires $\max(C_{in},C_{out})$ rounds.
The traversal speed of FedBRB is fixed as it only depends on the smallest client, while the traversal speed of FedRolex is closely related to the number of channels in the backbone model. If there are lots of channels in some layer, the speed of FedBRB will be much faster than FedRolex's.

\section{Experiment}
\subsection{Experiment Description}
\paragraph{Targeted Questions.}
In short, we conduct extensive experiments to answer the following questions:
\begin{itemize}
  \item \textbf{Q1}: Is FedBRB effective in small-to-large scenario?
  \item \textbf{Q2}: Is the efficacy of FedBRB attributed to the relatively large clients?
  \item \textbf{Q3}: Is directly removing relatively small clients feasible?
  \item \textbf{Q4}: Are contributions of FedBRB's components distinguishable?
\end{itemize}

\vspace{-12pt}
\paragraph{Model Heterogeneity.}
We follow the description of model size in HeteroFL \cite{diao2021heterofl}, using 5 letters \textit{a}-\textit{e} to represent 5 sizes $\sfrac{1}{1}$-$\sfrac{1}{16}$.
For example, \textit{a} represents a sub-model with equal size to the global model and \textit{b} means $\sfrac{1}{2}$ size.
The distribution of client sizes can be represented as a string where the number after a letter indicate sampling weight. For example, the string $a0$-$b1$-$c1$-$d1$-$e1$ represents a FL system containing $\sfrac{1}{2}$, $\sfrac{1}{4}$, $\sfrac{1}{8}$ and $\sfrac{1}{16}$ clients with equal possibility (25\% each) and no $\sfrac{1}{1}$ sub-models.
Note that $a0$ is the key signal of small-to-large scenario, while $a1$ represents the traditional device-heterogeneity FL situation.

\vspace{-12pt}
\paragraph{Dataset and Data Heterogeneity.}
We evaluate the performance of FedBRB on datasets including CIFAR-10 \cite{krizhevsky2009learning}, CIFAR-100 \cite{krizhevsky2009learning} and MNIST \cite{deng2012mnist}.
For data-heterogeneity, we follow the non-iid setting in HeteroFL and FedRolex, whereby each client's local data contains only examples of $L$ labels in all labeled categories. By setting specific values of $L$, this approach can roughly approximate the Dirichlet distribution $DirK(\alpha)$. For CIFAR-10 and MNIST, $L=2$ and $L=5$ corresponds to $DirK(0.1)$ and $DirK(0.5)$. And for CIFAR-100, $L=20$ and $L=50$ corresponds to $DirK(0.1)$ and $DirK(0.5)$.
Due to limitations in paper length, the experimental results for CIFAR-100 and MNIST are provided in the \underline{supplementary material}.

\vspace{-12pt}
\paragraph{Models and Baselines.}
We adopt PreResNet18 \cite{he2016deep}, static batch normalization \cite{diao2021heterofl} and scalar modules \cite{diao2021heterofl} following the settings of FedRolex.
We compare FedBRB to current state-of-the-art device-heterogeneity FL model partitioning techniques including HeteroFL \cite{diao2021heterofl} and FedRolex \cite{alam2022fedrolex} under identical learning rate, max global rounds, local training rounds, and learning rate decay schedule. We set 100 clients with 10\% selected ratio for all datasets.

\begin{table*}[t]
  \centering
  \scalebox{0.8}{
    \begin{tabular}{cccccccccc}
      \toprule
      \textbf{CIFAR-10}       & \multicolumn{3}{c}{\textbf{non-iid-2}} & \multicolumn{3}{c}{\textbf{non-iid-5}} & \multicolumn{3}{c}{\textbf{iid}}                                                                                                                                                                                                                                           \\
      \cmidrule(r){1-1} \cmidrule(r){2-4}  \cmidrule(r){5-7} \cmidrule(r){8-10}
      \textbf{fixed}          & \textbf{HeteroFL}                      & \textbf{FedRolex}                      & {\underline{\textbf{FedBRB}}}                & \textbf{HeteroFL}             & \textbf{FedRolex}             & {\underline{\textbf{FedBRB}}}                & \textbf{HeteroFL}             & \textbf{FedRolex}             & {\underline{\textbf{FedBRB}}}                \\ \midrule
      \textbf{a0-e1}          & \cellcolor[HTML]{C0C0C0}19.38          & \cellcolor[HTML]{C0C0C0}38.05          & \textbf{\underline{53.51}} ($^\uparrow$15.5) & \cellcolor[HTML]{C0C0C0}25.88 & \cellcolor[HTML]{C0C0C0}44.46 & \textbf{\underline{62.75}} ($^\uparrow$19.2) & \cellcolor[HTML]{C0C0C0}24.20 & \cellcolor[HTML]{C0C0C0}52.67 & \textbf{\underline{61.62}} ($^\uparrow$8.95) \\
      \textbf{a0-d1}          & \cellcolor[HTML]{C0C0C0}22.82          & \cellcolor[HTML]{C0C0C0}47.15          & \textbf{59.37} ($^\uparrow$12.2)             & \cellcolor[HTML]{C0C0C0}27.81 & 63.09                         & \textbf{65.52} ($^\uparrow$2.96)             & \cellcolor[HTML]{C0C0C0}25.41 & \textbf{70.67}                & 67.96 ($^\downarrow$2.71)                    \\
      \textbf{a0-c1}          & \cellcolor[HTML]{C0C0C0}23.57          & 55.73                                  & \textbf{68.87} ($^\uparrow$13.1)             & \cellcolor[HTML]{C0C0C0}29.92 & 71.51                         & \textbf{77.37} ($^\uparrow$6.05)             & \cellcolor[HTML]{C0C0C0}28.26 & 75.94                         & \textbf{78.12} ($^\uparrow$2.18)             \\
      \textbf{a0-b1}          & \cellcolor[HTML]{C0C0C0}30.52          & 65.41                                  & \textbf{73.72} ($^\uparrow$8.31)             & \cellcolor[HTML]{C0C0C0}42.30 & 81.87                         & \textbf{87.80} ($^\uparrow$5.93)             & \cellcolor[HTML]{C0C0C0}40.27 & 85.30                         & \textbf{90.04} ($^\uparrow$4.74)             \\ \midrule
      \textbf{a0-d1-e1}       & \cellcolor[HTML]{C0C0C0}24.65          & \cellcolor[HTML]{C0C0C0}41.92          & \textbf{50.65} ($^\uparrow$8.73)             & \cellcolor[HTML]{C0C0C0}31.54 & \textbf{59.97}                & 57.71 ($^\downarrow$2.26)                    & \cellcolor[HTML]{C0C0C0}30.95 & 68.81                         & \textbf{71.10} ($^\uparrow$2.29)             \\
      \textbf{a0-c1-e1}       & \cellcolor[HTML]{C0C0C0}25.98          & \cellcolor[HTML]{C0C0C0}46.55          & \textbf{53.74} ($^\uparrow$7.19)             & \cellcolor[HTML]{C0C0C0}36.79 & 69.58                         & \textbf{73.67} ($^\uparrow$4.09)             & \cellcolor[HTML]{C0C0C0}34.28 & 74.79                         & \textbf{81.26} ($^\uparrow$6.47)             \\
      \textbf{a0-b1-e1}       & \cellcolor[HTML]{C0C0C0}29.73          & \cellcolor[HTML]{C0C0C0}46.78          & \textbf{58.47} ($^\uparrow$11.7)             & \cellcolor[HTML]{C0C0C0}45.34 & 77.80                         & \textbf{82.54} ($^\uparrow$4.74)             & \cellcolor[HTML]{C0C0C0}48.59 & 81.77                         & \textbf{86.62} ($^\uparrow$4.85)             \\
      \textbf{a0-c1-d1}       & \cellcolor[HTML]{C0C0C0}26.89          & \cellcolor[HTML]{C0C0C0}48.39          & \textbf{60.70} ($^\uparrow$12.3)             & \cellcolor[HTML]{C0C0C0}32.99 & 70.91                         & \textbf{76.42} ($^\uparrow$5.51)             & \cellcolor[HTML]{C0C0C0}32.36 & 75.05                         & \textbf{76.80} ($^\uparrow$1.75)             \\
      \textbf{a0-b1-d1}       & \cellcolor[HTML]{C0C0C0}31.29          & \cellcolor[HTML]{C0C0C0}48.49          & \textbf{63.88} ($^\uparrow$15.4)             & \cellcolor[HTML]{C0C0C0}44.67 & 79.06                         & \textbf{82.71} ($^\uparrow$3.65)             & \cellcolor[HTML]{C0C0C0}46.45 & \textbf{84.28}                & 83.18 ($^\downarrow$1.10)                    \\
      \textbf{a0-b1-c1}       & \cellcolor[HTML]{C0C0C0}31.75          & 57.86                                  & \textbf{67.22} ($^\uparrow$9.36)             & \cellcolor[HTML]{C0C0C0}45.72 & 80.51                         & \textbf{82.21} ($^\uparrow$1.70)             & \cellcolor[HTML]{C0C0C0}49.28 & 83.15                         & \textbf{85.42} ($^\uparrow$2.27)             \\ \midrule
      \textbf{a0-c1-d1-e1}    & \cellcolor[HTML]{C0C0C0}27.04          & \cellcolor[HTML]{C0C0C0}44.65          & \textbf{54.85} ($^\uparrow$10.2)             & \cellcolor[HTML]{C0C0C0}38.76 & 69.15                         & \textbf{69.96} ($^\uparrow$0.81)             & \cellcolor[HTML]{C0C0C0}38.30 & 74.62                         & \textbf{76.96} ($^\uparrow$2.34)             \\ \midrule
      \textbf{a0-b1-c1-d1-e1} & \cellcolor[HTML]{C0C0C0}30.81          & \cellcolor[HTML]{C0C0C0}44.78          & \textbf{59.75} ($^\uparrow$15.0)             & \cellcolor[HTML]{C0C0C0}48.73 & 75.46                         & \textbf{79.38} ($^\uparrow$3.92)             & \cellcolor[HTML]{C0C0C0}49.96 & 80.31                         & \textbf{82.00} ($^\uparrow$1.69)             \\ \bottomrule
    \end{tabular}
  }
  \caption{(\textbf{Q1}) Global model accuracy comparison between FedBRB, HeteroFL and FedRolex on \textbf{CIFAR-10} under \textbf{fixed} setting.}
  \label{CIFAR10_fixed}
\end{table*}

\vspace{-12pt}
\paragraph{Dynamic and fixed settings.}
We follow HeteroFL's dynamic and fixed settings for size selection. In dynamic setting, clients can vary model size each round based on current available resources.
In fixed setting, clients select a model size once initially which remains fixed across rounds.
The fixed setting offers more predictability and stability, while the dynamic setting enables better adaptability to changing local environments.

\subsection{Main Experiments (Q1)}
\cref{CIFAR10_dynamic} and
\cref{CIFAR10_fixed}
show global performance comparisons in small-to-large scenario ($a0$) under dynamic/fixed settings.

\vspace{-12pt}
\paragraph{Dynamic setting training.}
From \cref{CIFAR10_dynamic} we can obtain following observations.
(1) FedBRB outperforms HeteroFL and FedRolex in the majority of cases, with some metrics improving by up to 20\% over FedRolex. This demonstrates FedBRB's superior performance in the small-to-large scenario, greatly alleviating the drastic degradation caused by lacking full-size sub-models.
(2) The participation of smaller clients can better reflect the superiority of FedBRB which is usually most pronounced under the $a0$-$e1$ condition.
This is because for HeteroFL and FedRolex, the smaller the participating local models, the larger their blank parameter space.
When local models of size $b$ participate in training, the blank parameter space ratio is only $\sfrac{3}{4}$ for HeteroFL, and only $\sfrac{1}{2}$ for FedRolex. However, when only $e$-level sub-models participate, the blank spaces are $\sfrac{255}{256}$ and $\sfrac{15}{16}$ for the two methods respectively which are much larger.
(3) FedBRB's performance improvement varies under different data distributions. Compared to FedRolex, FedBRB shows larger improvements under non-iid settings, while the increase is slightly smaller under iid setting. This is because with iid data, the importance of small local models decreases since similar data can be learned from other larger clients. However, for non-iid settings, the data owned by each client may be unique and indispensable, which highlights FedBRB's promotion of small models. Moreover, as data-heterogeneity increases, FedBRB's advantages become more significant.
(4) FedBRB using only minimal local models (\ie, \textbf{underlined} FedBRB results under $a0$-$e1$) can even surpass baselines using larger local models (\ie, results of comparison method with \textbf{shadow}). This shows that FedBRB not only outperforms FedRolex and HeteroFL with the same client computing resources, but can also sometimes defeat them with fewer resources.

\begin{figure}[t]
  \centering
  \setlength{\abovecaptionskip}{5pt}
  \setlength{\belowcaptionskip}{-10pt}
  \includegraphics[width=8cm]{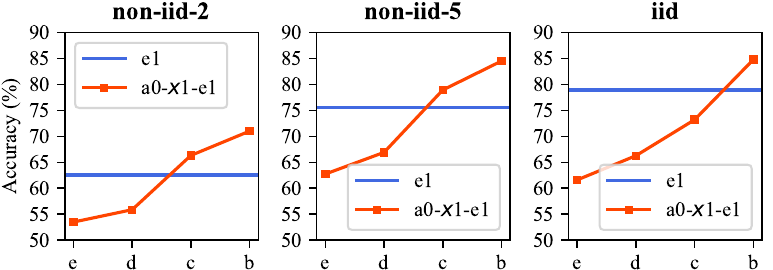}
  \caption{(\textbf{Q2}) Performance of $e$-level full model size training (blue) and FedBRB small-to-large scenario (red) on CIFAR-10 under dynamic setting. As the value of x in a0-$\underline{x}$1-e1 increases from e to b, the performance of FedBRB gradually surpasses the blue. This demonstrates the vital role of relatively large local models in FedBRB.}
  \label{Q2}
\end{figure}

\vspace{-12pt}
\paragraph{Fixed setting training.}
\cref{CIFAR10_fixed}
show the performances comparisons under fixed setting. Different from dynamic setting, fixed setting means clients cannot switch to a larger model in the next round to have greater impact on global model. Once clients initially choose smaller model sizes, the effects of their data will be limited throughout the process. Therefore, we can see that the performance of all models decreases under fixed setting compared to dynamic setting, especially under non-iid data settings.
More importantly, we also obtain similar observations as under the dynamic setting, including that FedBRB outperforms HeteroFL and FedRolex in the vast majority of cases. This means that whether it is small devices with dynamically changing computations or static computations, they can all play a greater role under the FedBRB framework.

\subsection{Compared to the full size training (Q2)}
\cref{Q2} shows results for traditional full model federated training (Hereinafter referred to as TF-training) \textit{vs} FedBRB small-to-large scenario on CIFAR-10 under dynamic setting.
Specifically, FedBRB makes better use of all clients with different computing power.
We can see that under non-iid-2 setting, TF-training's $e$-level complete global model gets the accuracy of 62.57\%, which would be gradually surpassed by FedBRB as the value of x in $a0$-$\underline{x}$1-$e1$ increases from $e$ to $b$.
This indicates that TF-training is inferior to FedBRB as not all clients can afford the bigger sub-models, nor all clients want to only maintain the $e$-level sub-models. The realistic device-heterogeneity scenario is that some clients can train larger models, while the remaining clients can only train smaller models. If TF-training is adopted, clients with stronger capabilities can only waste computing power to accommodate clients with weaker capabilities, thus failing to fully exerting their true effectiveness.
In contrast, FedBRB encourages each client to choose larger models according to their own computing power, and these larger sub-models will help other clients' smaller sub-models to pursue a better global performance.

One notable phenomenon is that the performance of the global $e$-level model trained by TF with $e1$ is better than the performance of FedBRB under $a0$-$e1$ distribution. This is a normal phenomenon, as the lossless integration of parameters of $e$-level sub-models into global large models without additional processes remains a significant research challenge.
And this phenomenon does not conflict with the advantages of FedBRB. FedBRB can fully leverage the computing power advantages of larger local models, accommodate various different local models in training, and promote mutual complementarity between models, without being constrained by unified local model sizes. Taking the above example again, FedBRB can choose the $a0$-$c1$-$e1$ distribution to achieve better performance than TF-training. This choice can give full play to the capabilities of each device and is more suitable for device-heterogeneity FL scenario.

\begin{figure}[t]
  \centering
  \setlength{\abovecaptionskip}{5pt}
  \setlength{\belowcaptionskip}{-10pt}
  \includegraphics[width=8cm]{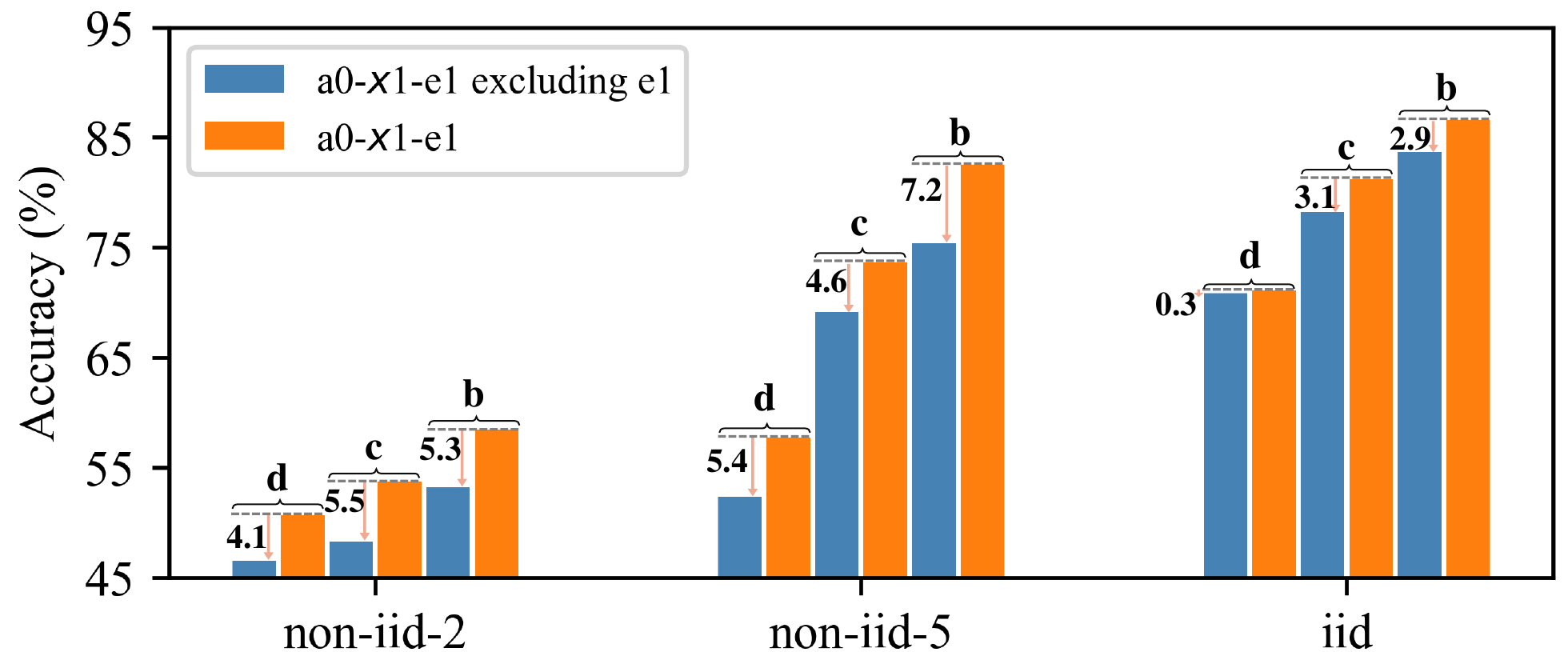}
  \caption{(\textbf{Q3}) Comparison experiments on CIFAR-10 under fixed setting between FedBRB (a0-$\underline{x}$1-e1) and FedBRB excluding small sub-models (a0-$\underline{x}$1-e1 excluding e1). The value of x in a0-$\underline{x}$1-e1 can be set to $d$, $c$ and $b$.}
  \label{Q3}
\end{figure}
\vspace{-3pt}

\subsection{Experiments excluding small sub-models (Q3)}
Since fusing local models of different sizes leads to performance loss, what if we drop the smaller clients and execute FedBRB directly on the larger clients? This could avoid fusion between the smaller models and the larger models. We conducted the following experiment to explore this question. Specifically, we tested on CIFAR10 with non-iid and iid data under fixed setting with different client distributions ranged from $a0$-$d1$-$e1$ to $a0$-$b1$-$e1$. The experiment procedure of FedBRB excluding small sub-models was to conduct FedBRB training process, but discard $e$-level sub-models during aggregation in each global round. Besides, the fixed setting allows us to clearly isolate and see the effect of removing small models. This removal is of little significance under dynamic setting, so we did not conduct experiments under dynamic setting.

\cref{Q3} shows results of the experiment.
We can see that FedBRB comprehensively outperforms the method excluding $e1$. It is because that discarding small models leads to decreased amounts of training data. Especially under non-iid settings, the performance degradation caused by the loss of high-value data is more serious.

\begin{table}[t]
  \centering
  \setlength{\abovecaptionskip}{5pt}
  \setlength{\belowcaptionskip}{-5pt}
  \scalebox{0.5}{
    \begin{tabular}{ccccccc}
      \toprule
      \textbf{CIFAR-10} & \multicolumn{2}{c}{\textbf{non-iid-2}} & \multicolumn{2}{c}{\textbf{non-iid-5}} & \multicolumn{2}{c}{\textbf{iid}}                                                              \\
      \cmidrule(r){1-1} \cmidrule(r){2-3}  \cmidrule(r){4-5} \cmidrule(r){6-7}
      \textbf{dynamic}  & \textbf{FedBRB w/o wB}                 & \textbf{FedBRB}                        & \textbf{FedBRB w/o wB}           & \textbf{FedBRB} & \textbf{FedBRB w/o wB} & \textbf{FedBRB} \\ \midrule
      a0-e1             & 51.95                                  & \textbf{53.51}                         & 58.36                            & \textbf{62.75}  & 61.19                  & \textbf{61.62}  \\
      a0-d1             & 56.51                                  & \textbf{59.37}                         & 63.27                            & \textbf{65.52}  & 66.13                  & \textbf{67.96}  \\
      a0-c1             & 66.39                                  & \textbf{68.87}                         & 76.01                            & \textbf{77.37}  & 76.75                  & \textbf{78.12}  \\
      a0-b1             & 72.16                                  & \textbf{73.72}                         & \textbf{88.13}                   & 87.80           & 89.18                  & \textbf{90.04}  \\ \midrule
      a0-d1-e1          & 53.68                                  & \textbf{55.89}                         & 62.41                            & \textbf{66.92}  & 63.97                  & \textbf{66.28}  \\
      a0-c1-e1          & 63.10                                  & \textbf{66.33}                         & 75.28                            & \textbf{78.92}  & 72.07                  & \textbf{73.21}  \\
      a0-b1-e1          & 68.43                                  & \textbf{71.01}                         & 80.96                            & \textbf{84.49}  & 82.94                  & \textbf{84.77}  \\
      a0-c1-d1          & 64.06                                  & \textbf{67.01}                         & 74.32                            & \textbf{75.29}  & 76.00                  & \textbf{76.07}  \\
      a0-b1-d1          & 68.74                                  & \textbf{70.29}                         & \textbf{83.60}                   & 83.35           & 82.78                  & \textbf{83.15}  \\
      a0-b1-c1          & 71.82                                  & \textbf{73.21}                         & \textbf{85.24}                   & 84.81           & 83.49                  & \textbf{85.87}  \\ \midrule
      a0-c1-d1-e1       & 65.94                                  & \textbf{67.40}                         & 70.83                            & \textbf{71.70}  & 76.92                  & \textbf{78.17}  \\ \midrule
      a0-b1-c1-d1-e1    & 65.67                                  & \textbf{68.12}                         & 79.45                            & \textbf{82.31}  & 81.03                  & \textbf{83.69}  \\ \bottomrule
    \end{tabular}
  }
  \caption{(\textbf{Q4}) Ablation experiment on CIFAR-10 under dynamic setting. FedBRB w/o wB stands for the ablation variant of FedBRB that without \underline{w}eighed \underline{B}roadcast module.}
  \label{ablation}
\end{table}

\subsection{Ablation Experiment (Q4)}
We conducted ablation experiments to analyze the contribution of components in FedBRB.
Since Block-wise Rolling is the cornerstone of the method design, we only designed one ablation variant, namely FedBRB w/o wB (i.e., FedBRB without weighted Broadcast).

\cref{ablation} presents results of the ablation experiment on CIFAR-10 under dynamic setting, from which we can find that FedBRB's performance surpasses that of FedBRB w/o wB in the vast majority of cases, which shows that the weighed Broadcast module provides effective help to FedBRB. Specifically, the improvements from weighted broadcast module are more pronounced when smaller sub-models account for a high proportion. This is because broadcasting provides more contributions for information interaction in such cases, similar to the analysis for Q1.


\vspace{-2pt}
\section{Conclusion}
In this paper, we clearly defines the \textit{small-to-large scenario} and elucidates its importance in device-heterogeneity federated learning for training large-scale global models. And we proposes an effective solution called FedBRB tailored for it. By employing the \textit{Block-wise Rolling} and \textit{weighted Broadcast}, FedBRB solves the problem of incomplete parameter coverage in FedRolex. Extensive experiments demonstrate FedBRB significantly improves the global model's performance in small-to-large scenario, which is of great significance for resource-constrained institutions nowadays.
{
    \small
    \bibliographystyle{ieeenat_fullname}
    \bibliography{main}

\begin{thebibliography}{34}
\providecommand{\natexlab}[1]{#1}
\providecommand{\url}[1]{\texttt{#1}}
\expandafter\ifx\csname urlstyle\endcsname\relax
  \providecommand{\doi}[1]{doi: #1}\else
  \providecommand{\doi}{doi: \begingroup \urlstyle{rm}\Url}\fi

\bibitem[Abdelmoniem and Canini(2021)]{abdelmoniem2021towards}
Ahmed~M Abdelmoniem and Marco Canini.
\newblock Towards mitigating device heterogeneity in federated learning via
  adaptive model quantization.
\newblock In \emph{Proceedings of the 1st Workshop on Machine Learning and
  Systems}, pages 96--103, 2021.

\bibitem[Alam et~al.(2022)Alam, Liu, Yan, and Zhang]{alam2022fedrolex}
Samiul Alam, Luyang Liu, Ming Yan, and Mi Zhang.
\newblock Fedrolex: Model-heterogeneous federated learning with rolling
  sub-model extraction.
\newblock \emph{Advances in Neural Information Processing Systems},
  35:\penalty0 29677--29690, 2022.

\bibitem[Aledhari et~al.(2020)Aledhari, Razzak, Parizi, and
  Saeed]{aledhari2020federated}
Mohammed Aledhari, Rehma Razzak, Reza~M Parizi, and Fahad Saeed.
\newblock Federated learning: A survey on enabling technologies, protocols, and
  applications.
\newblock \emph{IEEE Access}, 8:\penalty0 140699--140725, 2020.

\bibitem[Caldas et~al.(2018)Caldas, Kone{\v{c}}ny, McMahan, and
  Talwalkar]{caldas2018expanding}
Sebastian Caldas, Jakub Kone{\v{c}}ny, H~Brendan McMahan, and Ameet Talwalkar.
\newblock Expanding the reach of federated learning by reducing client resource
  requirements.
\newblock \emph{arXiv preprint arXiv:1812.07210}, 2018.

\bibitem[Chen et~al.(2022)Chen, Wang, Zhou, Chen, Xiao, and Lin]{chen2022cfl}
Qian Chen, Zilong Wang, Yilin Zhou, Jiawei Chen, Dan Xiao, and Xiaodong Lin.
\newblock Cfl: Cluster federated learning in large-scale peer-to-peer networks.
\newblock In \emph{International Conference on Information Security}, pages
  464--472. Springer, 2022.

\bibitem[Cho et~al.(2022)Cho, Manoel, Joshi, Sim, and
  Dimitriadis]{cho2022heterogeneous}
Yae~Jee Cho, Andre Manoel, Gauri Joshi, Robert Sim, and Dimitrios Dimitriadis.
\newblock Heterogeneous ensemble knowledge transfer for training large models
  in federated learning.
\newblock \emph{arXiv preprint arXiv:2204.12703}, 2022.

\bibitem[Dash et~al.(2022)Dash, Sharma, and Ali]{dash2022federated}
Bibhu Dash, Pawankumar Sharma, and Azad Ali.
\newblock Federated learning for privacy-preserving: A review of pii data
  analysis in fintech.
\newblock \emph{International Journal of Software Engineering \& Applications
  (IJSEA)}, 13\penalty0 (4), 2022.

\bibitem[Deng(2012)]{deng2012mnist}
Li Deng.
\newblock The mnist database of handwritten digit images for machine learning
  research.
\newblock \emph{IEEE Signal Processing Magazine}, 29\penalty0 (6):\penalty0
  141--142, 2012.

\bibitem[Diao et~al.(2021)Diao, Ding, and Tarokh]{diao2021heterofl}
Enmao Diao, Jie Ding, and Vahid Tarokh.
\newblock Hetero{\{}fl{\}}: Computation and communication efficient federated
  learning for heterogeneous clients.
\newblock In \emph{International Conference on Learning Representations}, 2021.

\bibitem[Fan et~al.(2023)Fan, Kang, Ma, Chen, Wei, Fan, and Yang]{fan2023fate}
Tao Fan, Yan Kang, Guoqiang Ma, Weijing Chen, Wenbin Wei, Lixin Fan, and Qiang
  Yang.
\newblock Fate-llm: A industrial grade federated learning framework for large
  language models.
\newblock \emph{arXiv preprint arXiv:2310.10049}, 2023.

\bibitem[Golub and Reinsch(1971)]{golub1971singular}
Gene~H Golub and Christian Reinsch.
\newblock Singular value decomposition and least squares solutions.
\newblock In \emph{Handbook for Automatic Computation: Volume II: Linear
  Algebra}, pages 134--151. Springer, 1971.

\bibitem[Gou et~al.(2021)Gou, Yu, Maybank, and Tao]{gou2021knowledge}
Jianping Gou, Baosheng Yu, Stephen~J Maybank, and Dacheng Tao.
\newblock Knowledge distillation: A survey.
\newblock \emph{International Journal of Computer Vision}, 129:\penalty0
  1789--1819, 2021.

\bibitem[Guo et~al.(2023)Guo, Chen, Wu, and Wang]{guo2023aigc}
Danhuai Guo, Huixuan Chen, Ruoling Wu, and Yangang Wang.
\newblock Aigc challenges and opportunities related to public safety: A case
  study of chatgpt.
\newblock \emph{Journal of Safety Science and Resilience}, 2023.

\bibitem[Hard et~al.(2018)Hard, Rao, Mathews, Ramaswamy, Beaufays, Augenstein,
  Eichner, Kiddon, and Ramage]{hard2018federated}
Andrew Hard, Kanishka Rao, Rajiv Mathews, Swaroop Ramaswamy, Fran{\c{c}}oise
  Beaufays, Sean Augenstein, Hubert Eichner, Chlo{\'e} Kiddon, and Daniel
  Ramage.
\newblock Federated learning for mobile keyboard prediction.
\newblock \emph{arXiv preprint arXiv:1811.03604}, 2018.

\bibitem[He et~al.(2020)He, Annavaram, and Avestimehr]{he2020group}
Chaoyang He, Murali Annavaram, and Salman Avestimehr.
\newblock Group knowledge transfer: Federated learning of large cnns at the
  edge.
\newblock \emph{Advances in Neural Information Processing Systems},
  33:\penalty0 14068--14080, 2020.

\bibitem[He et~al.(2016)He, Zhang, Ren, and Sun]{he2016deep}
Kaiming He, Xiangyu Zhang, Shaoqing Ren, and Jian Sun.
\newblock Deep residual learning for image recognition.
\newblock In \emph{Proceedings of the IEEE conference on computer vision and
  pattern recognition}, pages 770--778, 2016.

\bibitem[Horvath et~al.(2021)Horvath, Laskaridis, Almeida, Leontiadis,
  Venieris, and Lane]{horvath2021fjord}
Samuel Horvath, Stefanos Laskaridis, Mario Almeida, Ilias Leontiadis, Stylianos
  Venieris, and Nicholas Lane.
\newblock Fjord: Fair and accurate federated learning under heterogeneous
  targets with ordered dropout.
\newblock \emph{Advances in Neural Information Processing Systems},
  34:\penalty0 12876--12889, 2021.

\bibitem[Itahara et~al.(2021)Itahara, Nishio, Koda, Morikura, and
  Yamamoto]{itahara2021distillation}
Sohei Itahara, Takayuki Nishio, Yusuke Koda, Masahiro Morikura, and Koji
  Yamamoto.
\newblock Distillation-based semi-supervised federated learning for
  communication-efficient collaborative training with non-iid private data.
\newblock \emph{IEEE Transactions on Mobile Computing}, 22\penalty0
  (1):\penalty0 191--205, 2021.

\bibitem[Krizhevsky et~al.(2009)Krizhevsky, Hinton,
  et~al.]{krizhevsky2009learning}
Alex Krizhevsky, Geoffrey Hinton, et~al.
\newblock Learning multiple layers of features from tiny images.
\newblock 2009.

\bibitem[Li et~al.(2021)Li, He, and Song]{li2021model}
Qinbin Li, Bingsheng He, and Dawn Song.
\newblock Model-contrastive federated learning.
\newblock In \emph{Proceedings of the IEEE/CVF conference on computer vision
  and pattern recognition}, pages 10713--10722, 2021.

\bibitem[Li et~al.(2020)Li, Sahu, Zaheer, Sanjabi, Talwalkar, and
  Smith]{li2020federated}
Tian Li, Anit~Kumar Sahu, Manzil Zaheer, Maziar Sanjabi, Ameet Talwalkar, and
  Virginia Smith.
\newblock Federated optimization in heterogeneous networks.
\newblock \emph{Proceedings of Machine learning and systems}, 2:\penalty0
  429--450, 2020.

\bibitem[Lin et~al.(2020)Lin, Kong, Stich, and Jaggi]{lin2020ensemble}
Tao Lin, Lingjing Kong, Sebastian~U Stich, and Martin Jaggi.
\newblock Ensemble distillation for robust model fusion in federated learning.
\newblock \emph{Advances in Neural Information Processing Systems},
  33:\penalty0 2351--2363, 2020.

\bibitem[McMahan et~al.(2017)McMahan, Moore, Ramage, Hampson, and
  y~Arcas]{mcmahan2017communication}
Brendan McMahan, Eider Moore, Daniel Ramage, Seth Hampson, and Blaise~Aguera y
  Arcas.
\newblock Communication-efficient learning of deep networks from decentralized
  data.
\newblock In \emph{Artificial intelligence and statistics}, pages 1273--1282.
  PMLR, 2017.

\bibitem[Mei et~al.(2022)Mei, Guo, Zhou, and Patel]{mei2022resource}
Yiqun Mei, Pengfei Guo, Mo Zhou, and Vishal Patel.
\newblock Resource-adaptive federated learning with all-in-one neural
  composition.
\newblock \emph{Advances in Neural Information Processing Systems},
  35:\penalty0 4270--4284, 2022.

\bibitem[Niu et~al.(2022)Niu, Prakash, Kundu, Lee, and
  Avestimehr]{niu2022federated}
Yue Niu, Saurav Prakash, Souvik Kundu, Sunwoo Lee, and Salman Avestimehr.
\newblock Federated learning of large models at the edge via principal
  sub-model training.
\newblock \emph{arXiv preprint arXiv:2208.13141}, 2022.

\bibitem[Panagakis et~al.(2021)Panagakis, Kossaifi, Chrysos, Oldfield,
  Nicolaou, Anandkumar, and Zafeiriou]{9420085}
Yannis Panagakis, Jean Kossaifi, Grigorios~G. Chrysos, James Oldfield,
  Mihalis~A. Nicolaou, Anima Anandkumar, and Stefanos Zafeiriou.
\newblock Tensor methods in computer vision and deep learning.
\newblock \emph{Proceedings of the IEEE}, 109\penalty0 (5):\penalty0 863--890,
  2021.

\bibitem[Park and Ko(2023)]{park2023fedhm}
JaeYeon Park and JeongGil Ko.
\newblock Fedhm: Practical federated learning for heterogeneous model
  deployments.
\newblock \emph{ICT Express}, 2023.

\bibitem[Ray(2023)]{ray2023chatgpt}
Partha~Pratim Ray.
\newblock Chatgpt: A comprehensive review on background, applications, key
  challenges, bias, ethics, limitations and future scope.
\newblock \emph{Internet of Things and Cyber-Physical Systems}, 2023.

\bibitem[Singh et~al.(2022)Singh, Singh, Singh, and Singh]{singh2022federated}
Pushpa Singh, Murari~Kumar Singh, Rajnesh Singh, and Narendra Singh.
\newblock Federated learning: Challenges, methods, and future directions.
\newblock In \emph{Federated Learning for IoT Applications}, pages 199--214.
  Springer, 2022.

\bibitem[Sohan and Basalamah(2023)]{sohan2023systematic}
Md~Fahimuzzman Sohan and Anas Basalamah.
\newblock A systematic review on federated learning in medical image analysis.
\newblock \emph{IEEE Access}, 2023.

\bibitem[Wang et~al.(2020)Wang, Sreenivasan, Rajput, Vishwakarma, Agarwal,
  Sohn, Lee, and Papailiopoulos]{wang2020attack}
Hongyi Wang, Kartik Sreenivasan, Shashank Rajput, Harit Vishwakarma, Saurabh
  Agarwal, Jy-yong Sohn, Kangwook Lee, and Dimitris Papailiopoulos.
\newblock Attack of the tails: Yes, you really can backdoor federated learning.
\newblock \emph{Advances in Neural Information Processing Systems},
  33:\penalty0 16070--16084, 2020.

\bibitem[Wolfrath et~al.(2022)Wolfrath, Sreekumar, Kumar, Wang, and
  Chandra]{wolfrath2022haccs}
Joel Wolfrath, Nikhil Sreekumar, Dhruv Kumar, Yuanli Wang, and Abhishek
  Chandra.
\newblock Haccs: heterogeneity-aware clustered client selection for accelerated
  federated learning.
\newblock In \emph{2022 IEEE International Parallel and Distributed Processing
  Symposium (IPDPS)}, pages 985--995. IEEE, 2022.

\bibitem[Zhang et~al.(2021)Zhang, Xie, Bai, Yu, Li, and Gao]{zhang2021survey}
Chen Zhang, Yu Xie, Hang Bai, Bin Yu, Weihong Li, and Yuan Gao.
\newblock A survey on federated learning.
\newblock \emph{Knowledge-Based Systems}, 216:\penalty0 106775, 2021.

\bibitem[Zhao et~al.(2018)Zhao, Li, Lai, Suda, Civin, and
  Chandra]{zhao2018federated}
Yue Zhao, Meng Li, Liangzhen Lai, Naveen Suda, Damon Civin, and Vikas Chandra.
\newblock Federated learning with non-iid data.
\newblock \emph{arXiv preprint arXiv:1806.00582}, 2018.

\end{thebibliography}
}

\clearpage
\setcounter{page}{1}
\setcounter{section}{0}
\maketitlesupplementary

In this supplementary material, we provide more experiment details and results to show the effectiveness of \textbf{FedBRB} (\underline{B}lock-wise \underline{R}olling and weighted \underline{B}roadcast).
\section{More Experiment Results}

\subsection{Dataset Statistics}
The statistics of the datasets are listed in \cref{statistics}. We preprocess the datasets using RandomCrop, RandomHorizontalFlip and Normalize, following FedRolex \cite{alam2022fedrolex} and HeteroFL \cite{diao2021heterofl}.

\begin{table}[ht]
    \centering
    \scalebox{0.53}{
        \begin{tabular}{cccccc}
            \toprule
            \textbf{Dataset}           & \textbf{Image Size}              & \textbf{Class Number} & \textbf{Train Examples} & \textbf{Test Examples}  & \textbf{Preprocessing} \\ \midrule
            \multirow{3}{*}{CIFAR-10}  & \multirow{3}{*}{{[}3, 32, 32{]}} & \multirow{3}{*}{10}   & \multirow{3}{*}{50,000} & \multirow{3}{*}{10,000} & RandomCrop             \\
                                       &                                  &                       &                         &                         & RandomHorizontalFlip   \\
                                       &                                  &                       &                         &                         & Normalize              \\ \midrule
            \multirow{3}{*}{CIFAR-100} & \multirow{3}{*}{{[}3, 32, 32{]}} & \multirow{3}{*}{100}  & \multirow{3}{*}{50,000} & \multirow{3}{*}{10,000} & RandomCrop             \\
                                       &                                  &                       &                         &                         & RandomHorizontalFlip   \\
                                       &                                  &                       &                         &                         & Normalize              \\ \midrule
            MNIST                      & {[}1, 28, 28{]}                  & 10                    & 60,000                  & 10,000                  & Normalize              \\ \bottomrule
        \end{tabular}
    }
    \caption{\label{statistics} Dataset statistics.}
\end{table}

\vspace{-12pt}
\subsection{Main Experiments on CIFAR-100 \& MNIST}
\cref{CIFAR100_dynamic}, \cref{CIFAR100_fixed}, \cref{MNIST_dynamic} and \cref{MNIST_fixed} show global performance comparisons in small-to-large scenario ($a0$) on CIFAR-100/MNIST under dynamic/fixed settings.

As evident in the tables, FedBRB continues to attain substantial performance improvements over the baseline methods on CIFAR-100 and MNIST, under both non-IID and IID data distributions. More in-depth analysis is available in the main text (experiment analysis of Q1, Page 6).

\subsection{Experimental Setup Details}
The experimental setup details are tabulated in \cref{setup}.
For fair comparison, all models in the same experiment group are set up with the same training configurations.
\begin{table}[h]
    \centering
    \scalebox{0.85}{
        \begin{tabular}{cccc}
            \toprule
                                  & \textbf{CIFAR-10} & \textbf{CIFAR-100} & \textbf{MNIST} \\ \midrule
            Local Epoch           & 5                 & 5                  & 5              \\
            Global Round          & 800               & 800                & 800            \\
            Batch Size            & 64                & 64                 & 64             \\
            Initial Learning Rate & 1e-1              & 1e-1               & 1e-1           \\
            Decay Interval        & 300               & 300                & 200            \\
            Decay Factor          & 0.25              & 0.25               & 0.1            \\
            Optimizer             & SGD               & SGD                & SGD            \\
            Momentum              & 0.9               & 0.9                & 0.9            \\
            Weight Decay          & 1e-3              & 1e-3               & 5e-4           \\ \bottomrule
        \end{tabular}
    }
    \caption{\label{setup} Experimental setup details of FedBRB, FedRolex and HeteroFL.}
\end{table}

\subsection{Algorithm}
\cref{algo} shows the procedure of FedBRB.
\SetKwBlock{FServerStep}{ServerStep($r$)}{ServerStepEnd}
\SetKwBlock{FServerStepRef}{ServerStep($r$)}{}
\SetKwBlock{FClientStep}{ClientStep($W^{(r)}_i$, $C_i$)}{ClientStepEnd}
\SetKwBlock{FClientStepRef}{ClientStep($W^{(r)}_i$, $C_i$)}{}
\SetKwBlock{FBegin}{\textbf{Begin}}{End}
\begin{algorithm}[h]
    \caption{FedBRB}
    \label{algo}
    \textbf{Input}: clients $C$, datasets $D$, sub-model size choices \{$\frac{1}{2^{[x]}}$\}, global training rounds $R$

    \textbf{Initialization}: local dataset $D_i$, global model $W^{(0)}$, clients' sub-model sizes, block shape

    \FBegin{

        \For{$r \gets 0$ \KwTo $R-1$}{

            \FServerStepRef{}

        }
    }

    \FServerStep{

    Sample client-subset $C^{(r)}$

    Pick starting block and generate $W^{(r)}_{\{C^{(r)}\}}$ using \textit{Block-wise Rolling}

    \For{$C_i \in C^{(r)}$}{

        Distribute $W^{(r)}_{C_i}$ to $C_i$

        Get $W^{(r)^{'}}_{C_i}$ from \FClientStepRef{}
    }

    Aggregate $W^{(r+1)}$ using \textit{Block Weighted Broadcast}

    }

    \FClientStep{
        Local train $W^{(r)}_i$ on $D_i$ using SGD

        Change/maintain sub-model's size for round $r+1$

        Return $W^{(r)^{'}}_{C_i}$
    }

\end{algorithm}

\begin{table*}[t]
    \centering
    \scalebox{0.8}{
        \begin{tabular}{cccccccccc}
            \toprule
            \textbf{CIFAR-100}      & \multicolumn{3}{c}{\textbf{non-iid-20}} & \multicolumn{3}{c}{\textbf{non-iid-50}} & \multicolumn{3}{c}{\textbf{iid}}                                                                                                                                                                                                    \\
            \cmidrule(r){1-1} \cmidrule(r){2-4}  \cmidrule(r){5-7} \cmidrule(r){8-10}
            \textbf{dynamic}        & \textbf{HeteroFL}                       & \textbf{FedRolex}                       & {\textbf{FedBRB}}                & \textbf{HeteroFL}             & \textbf{FedRolex}                      & {\textbf{FedBRB}}          & \textbf{HeteroFL}             & \textbf{FedRolex}             & {\textbf{FedBRB}}          \\ \midrule
            \textbf{a0-e1}          & \cellcolor[HTML]{C0C0C0}4.64            & \cellcolor[HTML]{C0C0C0}17.24           & \textbf{\underline{22.39}}       & \cellcolor[HTML]{C0C0C0}4.12  & \cellcolor[HTML]{C0C0C0}21.52          & \textbf{\underline{23.50}} & \cellcolor[HTML]{C0C0C0}4.19  & \cellcolor[HTML]{C0C0C0}20.05 & \textbf{\underline{23.37}} \\
            \textbf{a0-d1}          & \cellcolor[HTML]{C0C0C0}4.46            & \textbf{28.37}                          & 25.45                            & \cellcolor[HTML]{C0C0C0}3.67  & 25.37                                  & \textbf{25.88}             & \cellcolor[HTML]{C0C0C0}4.17  & 25.59                         & \textbf{26.10}             \\
            \textbf{a0-c1}          & \cellcolor[HTML]{C0C0C0}3.93            & 40.69                                   & \textbf{44.13}                   & \cellcolor[HTML]{C0C0C0}4.62  & 41.35                                  & \textbf{45.97}             & \cellcolor[HTML]{C0C0C0}4.51  & 36.87                         & \textbf{47.28}             \\
            \textbf{a0-b1}          & \cellcolor[HTML]{C0C0C0}8.71            & 52.97                                   & \textbf{57.20}                   & \cellcolor[HTML]{C0C0C0}9.54  & 56.84                                  & \textbf{58.12}             & \cellcolor[HTML]{C0C0C0}9.36  & 45.19                         & \textbf{46.90}             \\ \midrule
            \textbf{a0-d1-e1}       & \cellcolor[HTML]{C0C0C0}6.60            & \textbf{23.31}                          & 16.66                            & \cellcolor[HTML]{C0C0C0}6.10  & \cellcolor[HTML]{C0C0C0}\textbf{20.13} & 19.92                      & \cellcolor[HTML]{C0C0C0}5.57  & 24.69                         & \textbf{25.00}             \\
            \textbf{a0-c1-e1}       & \cellcolor[HTML]{C0C0C0}7.76            & 32.11                                   & \textbf{37.17}                   & \cellcolor[HTML]{C0C0C0}6.82  & 26.36                                  & \textbf{37.16}             & \cellcolor[HTML]{C0C0C0}6.21  & 34.67                         & \textbf{39.84}             \\
            \textbf{a0-b1-e1}       & \cellcolor[HTML]{C0C0C0}12.33           & 43.62                                   & \textbf{50.87}                   & \cellcolor[HTML]{C0C0C0}11.73 & 46.84                                  & \textbf{54.91}             & \cellcolor[HTML]{C0C0C0}10.16 & 34.80                         & \textbf{54.26}             \\
            \textbf{a0-c1-d1}       & \cellcolor[HTML]{C0C0C0}7.11            & 37.26                                   & \textbf{42.08}                   & \cellcolor[HTML]{C0C0C0}6.96  & 31.74                                  & \textbf{38.02}             & \cellcolor[HTML]{C0C0C0}6.85  & 38.98                         & \textbf{45.88}             \\
            \textbf{a0-b1-d1}       & \cellcolor[HTML]{C0C0C0}12.51           & 48.11                                   & \textbf{55.16}                   & \cellcolor[HTML]{C0C0C0}12.99 & 48.48                                  & \textbf{56.72}             & \cellcolor[HTML]{C0C0C0}10.73 & 50.30                         & \textbf{55.81}             \\
            \textbf{a0-b1-c1}       & \cellcolor[HTML]{C0C0C0}12.43           & 52.81                                   & \textbf{56.12}                   & \cellcolor[HTML]{C0C0C0}11.10 & 54.62                                  & \textbf{56.02}             & \cellcolor[HTML]{C0C0C0}11.19 & 52.63                         & \textbf{56.91}             \\ \midrule
            \textbf{a0-c1-d1-e1}    & \cellcolor[HTML]{C0C0C0}9.31            & 33.40                                   & \textbf{38.01}                   & \cellcolor[HTML]{C0C0C0}8.74  & 29.91                                  & \textbf{38.99}             & \cellcolor[HTML]{C0C0C0}7.65  & 32.84                         & \textbf{42.53}             \\ \midrule
            \textbf{a0-b1-c1-d1-e1} & \cellcolor[HTML]{C0C0C0}15.24           & 46.26                                   & \textbf{46.30}                   & \cellcolor[HTML]{C0C0C0}15.30 & 48.24                                  & \textbf{48.37}             & \cellcolor[HTML]{C0C0C0}12.58 & 45.77                         & \textbf{52.68}             \\ \bottomrule
        \end{tabular}
    }
    \caption{\label{CIFAR100_dynamic} (Q1) Global model accuracy comparison between FedBRB, HeteroFL and FedRolex on \textbf{CIFAR-100} under \textbf{dynamic} setting.}
\end{table*}

\begin{table*}[t]
    \centering
    \scalebox{0.8}{
        \begin{tabular}{cccccccccc}
            \toprule
            \textbf{CIFAR-100}      & \multicolumn{3}{c}{\textbf{non-iid-20}} & \multicolumn{3}{c}{\textbf{non-iid-50}} & \multicolumn{3}{c}{\textbf{iid}}                                                                                                                                                                                                    \\
            \cmidrule(r){1-1} \cmidrule(r){2-4}  \cmidrule(r){5-7} \cmidrule(r){8-10}
            \textbf{fixed}          & \textbf{HeteroFL}                       & \textbf{FedRolex}                       & {\textbf{FedBRB}}                & \textbf{HeteroFL}             & \textbf{FedRolex}                      & {\textbf{FedBRB}}          & \textbf{HeteroFL}             & \textbf{FedRolex}             & {\textbf{FedBRB}}          \\ \midrule
            \textbf{a0-e1}          & \cellcolor[HTML]{C0C0C0}4.64            & \cellcolor[HTML]{C0C0C0}17.24           & \textbf{\underline{22.39}}       & \cellcolor[HTML]{C0C0C0}4.12  & \cellcolor[HTML]{C0C0C0}21.52          & \textbf{\underline{23.50}} & \cellcolor[HTML]{C0C0C0}4.19  & \cellcolor[HTML]{C0C0C0}20.05 & \textbf{\underline{23.37}} \\
            \textbf{a0-d1}          & \cellcolor[HTML]{C0C0C0}4.46            & \textbf{28.37}                          & 25.45                            & \cellcolor[HTML]{C0C0C0}3.67  & 25.37                                  & \textbf{25.88}             & \cellcolor[HTML]{C0C0C0}4.17  & 25.59                         & \textbf{26.10}             \\
            \textbf{a0-c1}          & \cellcolor[HTML]{C0C0C0}3.93            & 40.69                                   & \textbf{44.13}                   & \cellcolor[HTML]{C0C0C0}4.62  & 41.35                                  & \textbf{45.97}             & \cellcolor[HTML]{C0C0C0}4.51  & 36.87                         & \textbf{47.28}             \\
            \textbf{a0-b1}          & \cellcolor[HTML]{C0C0C0}8.71            & 52.97                                   & \textbf{57.20}                   & \cellcolor[HTML]{C0C0C0}9.54  & 56.84                                  & \textbf{58.12}             & \cellcolor[HTML]{C0C0C0}9.36  & 45.19                         & \textbf{46.90}             \\ \midrule
            \textbf{a0-d1-e1}       & \cellcolor[HTML]{C0C0C0}5.53            & \cellcolor[HTML]{C0C0C0}\textbf{19.69}  & 15.84                            & \cellcolor[HTML]{C0C0C0}6.25  & \cellcolor[HTML]{C0C0C0}\textbf{19.00} & 17.55                      & \cellcolor[HTML]{C0C0C0}5.31  & \cellcolor[HTML]{C0C0C0}20.06 & \textbf{23.03}             \\
            \textbf{a0-c1-e1}       & \cellcolor[HTML]{C0C0C0}7.69            & 29.22                                   & \textbf{34.87}                   & \cellcolor[HTML]{C0C0C0}7.30  & 30.95                                  & \textbf{40.41}             & \cellcolor[HTML]{C0C0C0}7.64  & 33.33                         & \textbf{41.52}             \\
            \textbf{a0-b1-e1}       & \cellcolor[HTML]{C0C0C0}11.10           & 38.70                                   & \textbf{48.27}                   & \cellcolor[HTML]{C0C0C0}11.65 & 43.62                                  & \textbf{50.51}             & \cellcolor[HTML]{C0C0C0}11.28 & 45.21                         & \textbf{56.76}             \\
            \textbf{a0-c1-d1}       & \cellcolor[HTML]{C0C0C0}7.70            & 32.66                                   & \textbf{37.37}                   & \cellcolor[HTML]{C0C0C0}7.60  & 31.33                                  & \textbf{43.02}             & \cellcolor[HTML]{C0C0C0}7.56  & 35.02                         & \textbf{47.98}             \\
            \textbf{a0-b1-d1}       & \cellcolor[HTML]{C0C0C0}10.58           & 39.48                                   & \textbf{49.56}                   & \cellcolor[HTML]{C0C0C0}13.22 & 45.43                                  & \textbf{49.72}             & \cellcolor[HTML]{C0C0C0}12.00 & 48.93                         & \textbf{57.51}             \\
            \textbf{a0-b1-c1}       & \cellcolor[HTML]{C0C0C0}11.37           & 44.67                                   & \textbf{49.58}                   & \cellcolor[HTML]{C0C0C0}13.16 & 48.16                                  & \textbf{49.66}             & \cellcolor[HTML]{C0C0C0}11.62 & 53.21                         & \textbf{56.65}             \\ \midrule
            \textbf{a0-c1-d1-e1}    & \cellcolor[HTML]{C0C0C0}8.87            & 28.87                                   & \textbf{32.16}                   & \cellcolor[HTML]{C0C0C0}8.90  & \textbf{30.42}                         & 29.33                      & \cellcolor[HTML]{C0C0C0}7.63  & 33.24                         & \textbf{33.27}             \\ \midrule
            \textbf{a0-b1-c1-d1-e1} & \cellcolor[HTML]{C0C0C0}12.11           & 37.18                                   & \textbf{39.59}                   & \cellcolor[HTML]{C0C0C0}14.61 & \textbf{43.27}                         & 41.01                      & \cellcolor[HTML]{C0C0C0}13.34 & 47.87                         & \textbf{48.42}             \\ \bottomrule
        \end{tabular}
    }
    \caption{\label{CIFAR100_fixed} (Q1) Global model accuracy comparison between FedBRB, HeteroFL and FedRolex on \textbf{CIFAR-100} under \textbf{fixed} setting.}
\end{table*}

\begin{table*}[t]
    \centering
    \scalebox{0.8}{
        \begin{tabular}{cccccccccc}
            \toprule
            \textbf{MNIST}          & \multicolumn{3}{c}{\textbf{non-iid-2}} & \multicolumn{3}{c}{\textbf{non-iid-5}} & \multicolumn{3}{c}{\textbf{iid}}                                                                                                                                                                                  \\
            \cmidrule(r){1-1} \cmidrule(r){2-4}  \cmidrule(r){5-7} \cmidrule(r){8-10}
            \textbf{dynamic}        & \textbf{HeteroFL}                      & \textbf{FedRolex}                      & {\textbf{FedBRB}}                & \textbf{HeteroFL}             & \textbf{FedRolex}             & {\textbf{FedBRB}}          & \textbf{HeteroFL}             & \textbf{FedRolex}             & {\textbf{FedBRB}} \\ \midrule
            \textbf{a0-e1}          & \cellcolor[HTML]{C0C0C0}18.07          & \cellcolor[HTML]{C0C0C0}68.80          & \textbf{\underline{79.62}}       & \cellcolor[HTML]{C0C0C0}22.83 & \cellcolor[HTML]{C0C0C0}96.50 & \textbf{\underline{96.64}} & \cellcolor[HTML]{C0C0C0}28.78 & \cellcolor[HTML]{C0C0C0}97.00 & \textbf{97.22}    \\
            \textbf{a0-d1}          & \cellcolor[HTML]{C0C0C0}20.04          & \cellcolor[HTML]{C0C0C0}78.61          & \textbf{80.35}                   & \cellcolor[HTML]{C0C0C0}16.96 & {98.22}                       & \textbf{98.31}             & \cellcolor[HTML]{C0C0C0}30.08 & 98.42                         & \textbf{98.45}    \\
            \textbf{a0-c1}          & \cellcolor[HTML]{C0C0C0}21.67          & 83.76                                  & \textbf{87.12}                   & \cellcolor[HTML]{C0C0C0}19.11 & 98.91                         & \textbf{99.01}             & \cellcolor[HTML]{C0C0C0}47.11 & 98.86                         & \textbf{98.95}    \\
            \textbf{a0-b1}          & \cellcolor[HTML]{C0C0C0}46.94          & 94.95                                  & \textbf{95.25}                   & \cellcolor[HTML]{C0C0C0}77.08 & 99.53                         & \textbf{99.57}             & \cellcolor[HTML]{C0C0C0}96.20 & 99.35                         & \textbf{99.49}    \\ \midrule
            \textbf{a0-d1-e1}       & \cellcolor[HTML]{C0C0C0}13.36          & \cellcolor[HTML]{C0C0C0}{76.70}        & \textbf{83.52}                   & \cellcolor[HTML]{C0C0C0}34.23 & {98.41}                       & \textbf{98.56}             & \cellcolor[HTML]{C0C0C0}52.80 & 98.27                         & \textbf{98.74}    \\
            \textbf{a0-c1-e1}       & \cellcolor[HTML]{C0C0C0}18.72          & \cellcolor[HTML]{C0C0C0}71.50          & \textbf{92.48}                   & \cellcolor[HTML]{C0C0C0}42.53 & 99.00                         & \textbf{99.15}             & \cellcolor[HTML]{C0C0C0}77.38 & 98.85                         & \textbf{99.14}    \\
            \textbf{a0-b1-e1}       & \cellcolor[HTML]{C0C0C0}39.41          & 81.84                                  & \textbf{91.37}                   & \cellcolor[HTML]{C0C0C0}90.93 & 99.28                         & \textbf{99.41}             & \cellcolor[HTML]{C0C0C0}97.07 & \textbf{99.28}                & {99.18}           \\
            \textbf{a0-c1-d1}       & \cellcolor[HTML]{C0C0C0}15.02          & \cellcolor[HTML]{C0C0C0}74.79          & \textbf{92.52}                   & \cellcolor[HTML]{C0C0C0}39.18 & 99.02                         & \textbf{99.22}             & \cellcolor[HTML]{C0C0C0}72.09 & 98.86                         & \textbf{99.10}    \\
            \textbf{a0-b1-d1}       & \cellcolor[HTML]{C0C0C0}41.59          & \cellcolor[HTML]{C0C0C0}79.31          & \textbf{89.41}                   & \cellcolor[HTML]{C0C0C0}90.74 & 99.24                         & \textbf{99.47}             & 97.68                         & \textbf{99.36}                & {99.35}           \\
            \textbf{a0-b1-c1}       & \cellcolor[HTML]{C0C0C0}35.48          & 83.84                                  & \textbf{96.01}                   & \cellcolor[HTML]{C0C0C0}87.70 & 99.44                         & \textbf{99.54}             & 97.51                         & 99.34                         & \textbf{99.54}    \\ \midrule
            \textbf{a0-c1-d1-e1}    & \cellcolor[HTML]{C0C0C0}18.76          & \cellcolor[HTML]{C0C0C0}76.91          & \textbf{93.48}                   & \cellcolor[HTML]{C0C0C0}59.34 & 98.98                         & \textbf{98.99}             & \cellcolor[HTML]{C0C0C0}85.37 & 98.95                         & \textbf{99.03}    \\ \midrule
            \textbf{a0-b1-c1-d1-e1} & \cellcolor[HTML]{C0C0C0}33.39          & \cellcolor[HTML]{C0C0C0}75.64          & \textbf{92.50}                   & \cellcolor[HTML]{C0C0C0}90.47 & 99.26                         & \textbf{99.34}             & 97.41                         & 99.26                         & \textbf{99.27}    \\ \bottomrule
        \end{tabular}
    }
    \caption{\label{MNIST_dynamic} (Q1) Global model accuracy comparison between FedBRB, HeteroFL and FedRolex on \textbf{MNIST} under \textbf{dynamic} setting.}
\end{table*}

\begin{table*}[t]
    \centering
    \scalebox{0.8}{
        \begin{tabular}{cccccccccc}
            \toprule
            \textbf{MNIST}          & \multicolumn{3}{c}{\textbf{non-iid-2}} & \multicolumn{3}{c}{\textbf{non-iid-5}} & \multicolumn{3}{c}{\textbf{iid}}                                                                                                                                                                                  \\
            \cmidrule(r){1-1} \cmidrule(r){2-4}  \cmidrule(r){5-7} \cmidrule(r){8-10}
            \textbf{fixed}          & \textbf{HeteroFL}                      & \textbf{FedRolex}                      & {\textbf{FedBRB}}                & \textbf{HeteroFL}             & \textbf{FedRolex}             & {\textbf{FedBRB}}          & \textbf{HeteroFL}             & \textbf{FedRolex}             & {\textbf{FedBRB}} \\ \midrule
            \textbf{a0-e1}          & \cellcolor[HTML]{C0C0C0}18.07          & \cellcolor[HTML]{C0C0C0}68.80          & \textbf{\underline{79.62}}       & \cellcolor[HTML]{C0C0C0}22.83 & \cellcolor[HTML]{C0C0C0}96.50 & \textbf{\underline{96.64}} & \cellcolor[HTML]{C0C0C0}28.78 & \cellcolor[HTML]{C0C0C0}97.00 & \textbf{97.22}    \\
            \textbf{a0-d1}          & \cellcolor[HTML]{C0C0C0}20.04          & \cellcolor[HTML]{C0C0C0}78.61          & \textbf{80.35}                   & \cellcolor[HTML]{C0C0C0}16.96 & {98.22}                       & \textbf{98.31}             & \cellcolor[HTML]{C0C0C0}30.08 & 98.42                         & \textbf{98.45}    \\
            \textbf{a0-c1}          & \cellcolor[HTML]{C0C0C0}21.67          & 83.76                                  & \textbf{87.12}                   & \cellcolor[HTML]{C0C0C0}19.11 & 98.91                         & \textbf{99.01}             & \cellcolor[HTML]{C0C0C0}47.11 & 98.86                         & \textbf{98.95}    \\
            \textbf{a0-b1}          & \cellcolor[HTML]{C0C0C0}46.94          & 94.95                                  & \textbf{95.25}                   & \cellcolor[HTML]{C0C0C0}77.08 & 99.53                         & \textbf{99.57}             & \cellcolor[HTML]{C0C0C0}96.20 & 99.35                         & \textbf{99.49}    \\ \midrule
            \textbf{a0-d1-e1}       & \cellcolor[HTML]{C0C0C0}16.26          & \cellcolor[HTML]{C0C0C0}64.82          & \textbf{86.43}                   & \cellcolor[HTML]{C0C0C0}32.40 & \textbf{97.04}                & 97.01                      & \cellcolor[HTML]{C0C0C0}49.08 & 98.33                         & \textbf{96.13}    \\
            \textbf{a0-c1-e1}       & \cellcolor[HTML]{C0C0C0}21.04          & 80.74                                  & \textbf{88.93}                   & \cellcolor[HTML]{C0C0C0}43.51 & 98.70                         & \textbf{98.89}             & \cellcolor[HTML]{C0C0C0}80.07 & 98.81                         & \textbf{99.14}    \\
            \textbf{a0-b1-e1}       & \cellcolor[HTML]{C0C0C0}43.84          & 91.54                                  & \textbf{90.19}                   & \cellcolor[HTML]{C0C0C0}88.58 & 99.20                         & \textbf{99.43}             & 97.52                         & 99.27                         & \textbf{99.47}    \\
            \textbf{a0-c1-d1}       & \cellcolor[HTML]{C0C0C0}17.09          & 81.17                                  & \textbf{89.68}                   & \cellcolor[HTML]{C0C0C0}41.97 & 98.88                         & \textbf{99.21}             & \cellcolor[HTML]{C0C0C0}83.04 & 98.91                         & \textbf{99.07}    \\
            \textbf{a0-b1-d1}       & \cellcolor[HTML]{C0C0C0}36.49          & 90.19                                  & \textbf{90.36}                   & \cellcolor[HTML]{C0C0C0}88.06 & 99.33                         & \textbf{99.43}             & 97.47                         & 99.30                         & \textbf{99.42}    \\
            \textbf{a0-b1-c1}       & \cellcolor[HTML]{C0C0C0}31.79          & 91.18                                  & \textbf{91.22}                   & \cellcolor[HTML]{C0C0C0}86.44 & 99.32                         & \textbf{99.51}             & \cellcolor[HTML]{C0C0C0}92.98 & 99.41                         & \textbf{99.48}    \\ \midrule
            \textbf{a0-c1-d1-e1}    & \cellcolor[HTML]{C0C0C0}17.89          & 82.01                                  & \textbf{85.10}                   & \cellcolor[HTML]{C0C0C0}53.20 & 98.62                         & \textbf{98.87}             & \cellcolor[HTML]{C0C0C0}85.52 & \textbf{98.94}                & {98.65}           \\ \midrule
            \textbf{a0-b1-c1-d1-e1} & \cellcolor[HTML]{C0C0C0}41.56          & 90.49                                  & \textbf{90.32}                   & \cellcolor[HTML]{C0C0C0}91.41 & 99.07                         & \textbf{99.19}             & 97.81                         & 99.25                         & \textbf{99.38}    \\ \bottomrule
        \end{tabular}
    }
    \caption{\label{MNIST_fixed} (Q1) Global model accuracy comparison between FedBRB, HeteroFL and FedRolex on \textbf{MNIST} under \textbf{fixed} setting.}
\end{table*}


\end{document}